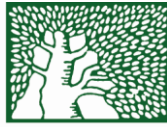

| Thesis for the degree<br>Doctor of Philosophy | עבודת גמר (תזה) לתואר<br>דוקטור לפילוסופיה |
|---|---|
| Submitted to the Scientific Council of the<br>Weizmann Institute of Science<br>Rehovot, Israel | מוגשת למועצה המדעית של<br>מכון ויצמן למדע<br>רחובות, ישראל |
| By<br>**Daniel Harari** | מאת<br>**דניאל הררי** |

**שימוש בתנועה ובהנחיה פנימית בזיהוי עצמים**

# Using Motion and Internal Supervision in Object Recognition

| Advisor:<br>**Prof. Shimon Ullman** | מנחה:<br>**פרופ' שמעון אולמן** |
|---|---|
| March 2012 | אדר תשע"ב |

# תקציר

תזה זו עוסקת בשני היבטים של זיהוי עצמים חזותי הקשורים זה בזה: האחד הוא השימוש במידע הנובע מתנועה בתמונה לצורך שיפור הזיהוי של עצמים על פני זמן, והשני הוא שימוש במנגנון של "הנחיה פנימית" לצורך סיוע בלמידה, מבלי להיעזר בהנחיה חיצונית. הקשר בין שני היבטים אלה בא לידי ביטוי במחקר הנוכחי, בכך שתנועה משמשת הן לצורך עקיבה וזיהוי עצמים גם כאשר תמונתם משתנה עם הזמן, והן לצורך יצירת מנגנון של הנחיה פנימית, באמצעות זיהוי של אירועים מיוחדים במרחב ובזמן, בפרט אירועים של תנועה אקטיבית. רוב העבודות הקיימות בנושא זיהוי אובייקטים עוסקות בתמונות ניחות גם במהלך הלמידה וגם במהלך הזיהוי. לעומת זאת, אנו מעוניינים בסצנה דינמית בה שינויים בתמונה עם הזמן תורמים מידע מועיל במרחב ובזמן, לצורך מיקוד תשומת לב אל אובייקט מסוים, לסגמנטציה של תנועה, ולהבנת התלת-מימד ואינטראקציות בין אובייקטים. אנו חוקרים את השימוש במקורות המידע הללו הן בתהליכי הלמידה והן בתהליכי הזיהוי.

בחלק הראשון של המחקר, אנו מראים כיצד תנועה יכולה לשמש לצורך זיהוי אדפטיבי של עצם בתנועה יחד עם כל החלקים המרכיבים אותו, תוך לימוד אוטומטי של הופעות חדשות של העצם וכל חלקיו.

בחלקו השני והעיקרי של המחקר אנו מפתחים שיטות לשימוש בסוגים מסוימים של תנועה בתמונה לפתרון שתי בעיות קשות בלמידה חזותית בלתי-מונחית: הראשונה היא למידה של זיהוי ידיים הן על פי המראה שלהם והן לפי ההקשר הסובב אותן, והשנייה היא למידה של זיהוי כיוון המבט של אדם בתמונה. אנו משתמשים במסקנותינו בחלק זה כדי להציע מודל עבור מספר היבטים של למידה של תינוקות אנושיים מהסביבה החזותית שלהם.

# Abstract

In this thesis we address two related aspects of visual object recognition: the use of motion information, and the use of internal supervision, to help unsupervised learning. These two aspects are inter-related in the current study, since image motion is used for internal supervision, via the detection of spatiotemporal events of active-motion and the use of tracking. Most current work in object recognition deals with static images during both learning and recognition. In contrast, we are interested in a dynamic scene where visual processes, such as detecting motion events and tracking, contribute spatiotemporal information, which is useful for object attention, motion segmentation, 3-D understanding and object interactions. We explore the use of these sources of information in both learning and recognition processes.

In the first part of the work, we demonstrate how motion can be used for adaptive detection of object-parts in dynamic environments, while automatically learning new object appearances and poses.

In the second and main part of the study we develop methods for using specific types of visual motion to solve two difficult problems in unsupervised visual learning: learning to recognize hands by their appearance and by context, and learning to extract direction of gaze. We use our conclusions in this part to propose a model for several aspects of learning by human infants from their visual environment.

# Acknowledgments

First and foremost I want to express my gratitude to my thesis advisor Shimon Ullman. Shimon has been a great mentor, an educator and a guide, whose wisdom does not fall behind his nobleness and modesty. The guidance, moral support and trust he provided throughout my studies, helped me keep on track and fulfill one of my best childhood dreams.

Second, I would like to thank Nimrod Dorfman, my research collaborator and friend, for the past three years. We have shared a joyful and inspiring joint work period, while exploring a fascinating research area, new to both of us.

I would also like to thank Leonid Karlinsky and Michael Dinerstein, my colleagues and collaborators, for a great research experience at the beginning of my studies, as well as for their professional and moral support.

I would like to thank Dr. Yahav Magen for her support and enlightening comments to my work, and also my aunt Dr. Rene Levi for providing me with an early inspiration of higher education.

Finally, I want to thank my closest family, my in laws, my sisters and my parents, David and Michele, for their moral support and help in everyday tasks, which made this study period feasible for me at this stage of my life.

Last but not least, I want to express my deepest gratitude to my darling wife, Shira Sheleg, and precious daughters, Maya and Mika, for their ever-lasting support. The persistent attention, claimed by my research work, comes always on their expense. Their unconditional love and care fulfill me with the mental strength and energy which I need the most.

To my dear parents, precious daughters, and beloved wife

*Live as if you were to die tomorrow. Learn as if you were to live forever.*

- M.K. Gandhi

# Table of Contents

# 1 Introduction

## 1.1 Overview of the study

Our world is a dynamic environment providing a dynamic visual input to the visual system. The dynamics of a scene introduce continuous changes of the appearance and spatial geometry of the visual content over time. These changes provide a rich source of information that can be used in many visual processes. Motion has been used in the past in various ways such as for directing attention, figure-ground segmentation, or the recovery of 3-D structure. In this study we focus on the use of motion information in recognition. We deal with two main aspects. One is the recognition of object that changes continuously in the visual input. The other is the use of visual motion as an internal teaching signal to help unsupervised learning. These two aspects are inter-related in the current work, since image motion is used for internal supervision, via the detection of spatiotemporal 'mover' events of active-motion and via the use of tracking. We refer to this unsupervised learning approach as internal supervision, since our learning processes are applied to natural dynamic visual input without external supervision.

In the first part of this work we consider a visual recognition system whose goal is to detect and interpret objects as they change over time in a dynamic environment. The dynamic changes are useful, since they provide cues for both segmentation and 3-D structure, but they are also challenging, as both the appearance and the relative positions of visual features may change over time. In this work we deal with a specific aspect of dynamic recognition. We assume that an initial object model successfully detects the object at some initial time $t_0$. Our goal is to continue to detect the object and all of its parts reliably as long as possible for later times $t > t_0$ while adapting the object model to the changing appearance and structure. We also show how motion information is used in this setting as internal supervision for an unsupervised learning process, which extends the initial object model to deal with a new set of viewing directions.

Specifically, we suggest an adaptive part-detection model for a dynamic environment, which is a natural extension of the existing part-detection methods for static images. The combination of a robust part detection model with the dynamics of the visual input yields a powerful recognition system, with capabilities beyond object tracking such as the automatic understanding of object appearance and sturcture in 3-D space. Our scheme combines the structure of the model at any time instance t with the optical flow between time t and (t + Δt). Using motion and spatiotemporal consistency, the model adapts online to dynamic changes of the observed object, by gradually updating both the appearance and structure of the object and its parts.

In the second and main part of this work we develop methods for using specific types of visual motion to provide an internal teaching signal for the learning of complex tasks such as recognizing hands and recovering direction of gaze in an unsupervised manner. Using a stream of unlabeled video sequences containing people performing everyday actions, the system learns to detect hands by their appearance and the surrounding context, and to extract people's direction of gaze. Hands are frequently engaged in motion, and their motion can provide a useful cue for acquiring hand-concepts. We, therefore, introduce in the model the detection of active motion, which we call 'mover' event, defined as the event of a moving image region causing a stationary region to move or change after contact. Our model detects 'mover' events and uses them as presumed hand examples to learn an initial appearance-based hand detector. Recognition capabilities of this hand detector are further extended by using the spatiotemporal continuity in tracking, and by learning to use body context. As hands may be detected based on either appearance or body context, we combine these two complementary detection methods iteratively in our model, to extend the range of appearances and poses used for learning by the algorithm. The performance of the hand detector improves rapidly during the first few iterations, approaching the performance of fully supervised training on the same data. Similarly, the use of detected 'mover' events as accurate teaching signals for learning to detect direction of gaze, allows our learning method to reach detection accuracy approaching adult human performance under similar test conditions. Finally, we use our conclusions in this part to propose a model for several aspects of learning by human infants from their visual environment.

In summary, we investigate in this study two inter-related components, motion and internal supervision, as fundamental mechanisms in the visual recognition process. We demonstrate their role in the recognition process as well as in the learning process, and develop several supervised and unsupervised computational methods for object recognition in video. We utilize well established statistical tools such as approximate nearest neighbors and kernel density estimation as underlying machinery for our models (Figure 1).

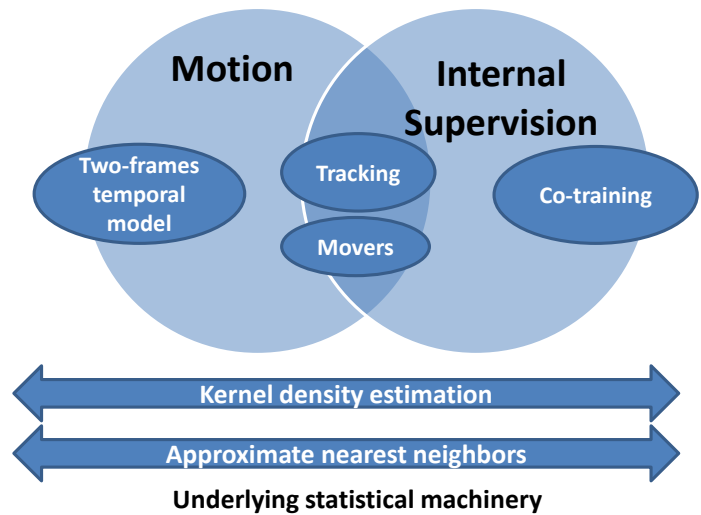

Figure 1: Outline of our study. We focus on two main inter-related themes: the use of visual motion and of internal supervision in object recognition. We develop several computation methods for object recognition in dynamic scenes, while using well established statistical tools such as KDE (kernel density estimation) and ANN (approximate nearest neighbors) throughout the work.

## 1.2 Outline of the dissertation

This dissertation consists of seven chapters. In chapter 2 we consider the task of adaptive object part detection in a dynamic environment. Using motion and spatiotemporal consistency, our suggested model adapts online to dynamic changes of the observed object, by gradually updating both appearance and structure of the object and its parts.

Chapter 3 presents our so-called 'mover' algorithm for unsupervised hand detection. We suggest that hands are learned based on the detection of special spatiotemporal 'mover' events, which are typical of hands and can be used as an internal 'tagging' of likely hand locations. Our testing shows that current computational methods for general object detection (such as saliency and informative image fragments), when applied to large training data, do not result by themselves, in automatically learning about hands.

In chapter 4 we suggest a method for the continuous improvement of recognition capabilities by co-training of two complementary classifiers using different cues in an iterative process. This unsupervised learning procedure integrates appearance-based object detection with context-based part detection (see Appendix A - The chains model for detecting parts by their context), while using each model detections as an internal supervision for training the other model.

Chapter 5 discusses our findings about an alternative developmental approach for learning hands, based on first-person perspective of own hands. Compared with the 'mover'-based detection, the own-hands detector performs inferiorly, and does not generalize to views of manipulating hands, which makes the own-hands detector inconsistent with data from infants' behavior during learning.

In chapter 6 we propose that 'mover' events can provide accurate teaching cues in the acquisition of another intriguing capacity in early visual perception – following another person's gaze based on head orientation. This skill, which begins to develop around 3-6 months of age, plays an important role in the development of communication and language. Our algorithm suggests how this learning might be accomplished, although cues for direction of gaze (head and eyes) can be subtle and difficult to extract and use.

Finally, chapter 7 summarizes the aspects presented in this study, and lists the main conclusions and contributions of the entire research work.

# 2 Adaptive Part Detection in a Dynamic Environment

The world around us is a dynamic environment, and a robust visual recognition system should therefore be able to detect and interpret objects as they change over time. The dynamic changes are useful, since they provide cues for both segmentation and 3D structure, but also challenging, as both the appearance and the relative positions of visual features may change over time.

In this work we deal with a specific aspect of dynamic recognition. We assume that an initial object model successfully detects the object at some time $t_0$, and the goal is to continue to detect the object and all of its parts reliably as long as possible for later times $t > t_0$ while adapting the object model to the changing appearance and structure, as illustrated in Figure 2. This process can also be used for learning, by extending the initial object model to deal with an extended set of viewing directions.

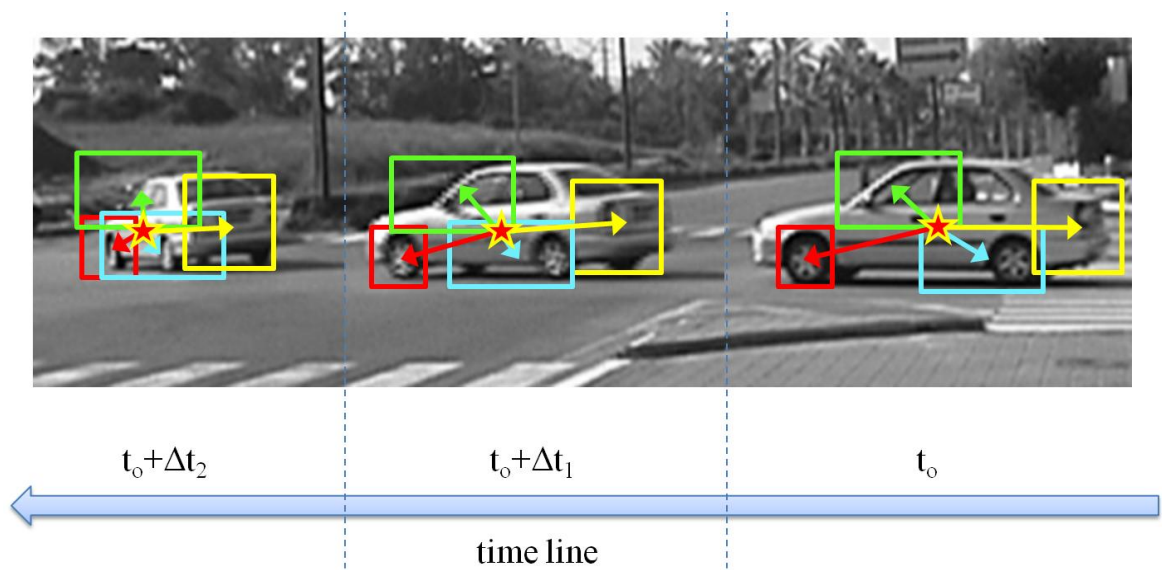

Figure 2: Illustration of object part detection in a dynamic visual input. The object and parts are detected initially at $t_0$. The goal is to continue to detect the parts for as long as possible, and extend the object model to deal with the novel view.

A main contribution of our approach is constructing the object model at $t > t_0$ by combining two sources of information: compatibility with the measured optical flow and similarity to the object model at an earlier time.

## *2.1* Related Work

We follow the paradigm of detecting and localizing objects by their constituent parts. Part-based object recognition has been successfully demonstrated in many recognition problems, mainly for detecting objects in static images (Agarwal *et al.*, 2004; Crandall *et al.*, 2005; Felzenszwalb and Huttenlocher, 2005; Fergus *et al.*, 2005; Epshtein and Ullman, 2007).

Object parts can be obtained manually (Felzenszwalb and Huttenlocher, 2005) or automatically (Ullman *et al.*, 2002; Agarwal *et al.*, 2004) during training from a set of sample images of the object. Each part is characterised by a visual appearance and by a spatial relation with other object parts. Felzenszwalb and Huttenlocher (Felzenszwalb and Huttenlocher, 2005) have suggested a pictorial structure representation using a collection of parts arranged in a deformable configuration. They model the appearance of each part separately, and represent the deformable configuration with spring-like connections between pairs of parts. Crandal et al. (Crandall *et al.*, 2005) have extended this approach by introducing a class of statistical models for part-based object recognition that are explicitly parameterized according to the degree of spatial structure they can represent. These models, called k-fans, provide a way of relating different spatial priors that have been used for recognizing generic classes of objects, including joint Gaussian models and tree-structured models.

A major contribution for the part-based recognition paradigm is accounted to Felzenszwalb *et al.* (Felzenszwalb *et al.*, 2010) by combining the powerful and robust histogram of gradients features (HOG) of Dalal and Triggs (Dalal and Triggs, 2005) into the part-based recognition framework. Their approach won both the 2008 and the 2009 PASCAL object detection challenge, using a star-structured part-based model defined by a set of object's "root" and parts HOG filters and associated deformation models.

A number of previous schemes have considered the application of object and part detection to dynamic input. Dalal *et al.* (Dalal *et al.*, 2006) have combined differential optical flow descriptors with their holistic object HOG descriptors for the task of detecting and tracking humans in video sequences. However, object parts are not represented in the model. A part-based human tracker was suggested by (Ioffe and Forsyth, 2001), using a mixture of trees to handle partial occlusions of some body parts. Temporal constraints are applied to enforce motion coherence across frames, and the object trajectory is inferred by maximizing the likelihood over the whole video sequence. Nevertheless, both the appearance and structure of the parts are learned during training and cannot be updated with the input dynamics.

A more recent approach to tracking by detection suggested by (Ramanan *et al.*, 2007), first learns specific appearance models of class objects from detections of a generic model applied to the visual input, and then tracks the objects . The inference is performed in an iterative manner for each object instance over the whole video sequence, while constraining the parts motion to bounded velocity. However, an online adaption of the model is not possible since the inference is performed over all the sequence frames.

Other approaches such as (Lim *et al.*, 2005; Kalal *et al.*, 2010) introduce online learning while tracking an object in order to improve the tracking capabilities. (Lim *et al.*, 2005) presented an online algorithm that incrementally learns and adapts a low dimensional eigenspace representation to reflect appearance changes of the target object. The tracking problem is formulated as a state inference problem within a MCMC framework, and a particle filter is incorporated for propagating sample distributions over time. However, object parts are not represented in the suggested model .

Recently, (Kalal *et al.*, 2010) have suggested a system for long-term tracking of a human face in unconstrained videos. The system is built on tracking-learning-detecting approach using an off-line trained generic detector and an online trained validation mechanism for pruning incorrect detections. A multi-view model of the target is automatically learned from a single frontal example and the unlabeled dynamic visual input. Nevertheless, object parts are not represented in the model, and the online validation mechanism is mainly application dependent.

In this work we present an adaptive parts detection model for a dynamic environment. The suggested scheme is a natural extension of the powerful and robust part detection paradigm of static images. The combination of a robust part detection model with the dynamics of the visual input yields with a powerful realistic recognition system, with capabilities beyond object tracking such as the automatic acquisition of object 3D understanding .

Our scheme combines the structure of the model at time t with the optical flow between time t and t + ∆t). Using motion and spatiotemporal consistency, the model adapts online to dynamic changes of the observed object, by gradually updating both appearance and structure of the object and its parts. Applying such a combination to obtain full part interpretation has not been done previously.

The rest of this chapter is organized as follows: in section 2.2 we describe the model and our probabilistic framework; in section 2.3 we present an experimental performance study; and in section 2.4 we discuss and conclude our insights from this work.

## *2.2* Methods

### 2.2.1 Overview

We consider an object recognition model applied to two consecutive image frames t and (t + ∆t) of a video sequence. Parts' interpretation (appearance and spatial position) at time (t + ∆t) is obtained by combining two sources: the model M(t) at time t, and the optical flow between the frames. The model is then updated to M(t + ∆t) to be used in the subsequent frame as shown in Figure 3. The models accumulated over time can also serve to construct an extended object model at different poses. This model can then be applied to static images of the object at those new poses.

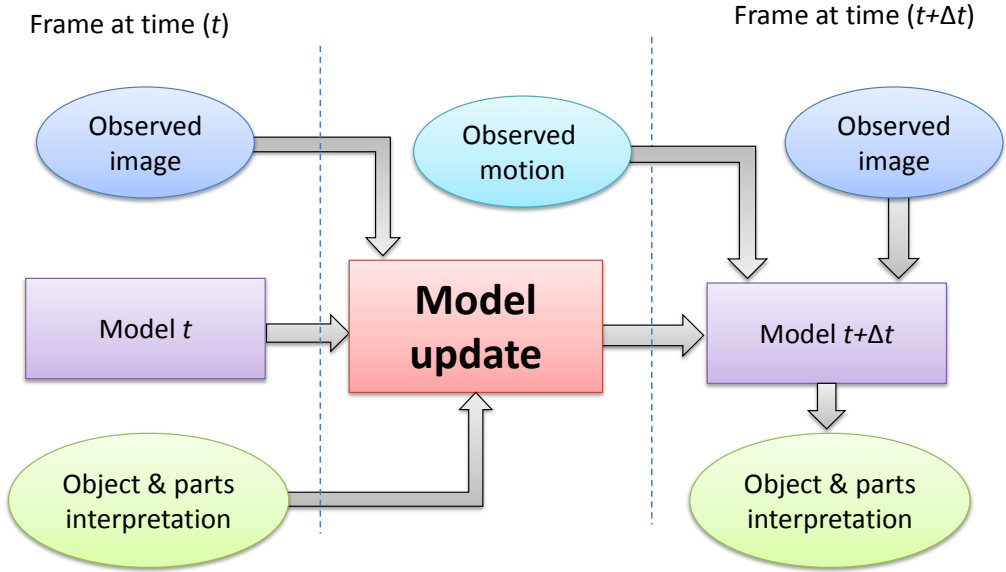

Figure 3: Temporal update scheme of the two-frame model. Parts' detection at time (t + Δt) is obtained by combining two sources: the model M(t) at time t, and the optical flow between the frames. The model is then updated to M(t + Δt) to be used in the subsequent frame. The model update includes adding new part appearances to the existing set of appearances, and updating the geometric structure parameters for successfully interpreted parts at time t.

When applied to video sequences, our model initially resembles a static, single-image parts detection model with a star-like geometric structure similar to (Crandall *et al.*, 2005; Fergus *et al.*, 2005; Epshtein and Ullman, 2007). Once the object is successfully detected at some time $t_0$, the model starts the adaption process as mentioned above. The updated models at each frame represent adapted instances of the initial detector learned by the input dynamics.

### 2.2.2 Probabilistic model

The initial static object detector is based on the representation of the object and its constituent parts following (Epshtein and Ullman, 2007). The appearances of parts and their geometric configurations are learned from positive static image samples which contain the class object and negative image samples which contain non-class background. The learning process may be in a fully supervised manner. However, we prefer the weakly supervised learning approach of object parts (Agarwal *et al.*, 2004) which is more realistic, and automatically select the parts from a large set of image fragments according to their mutual information with the object class (Vidal-Naquet and Ullman, 2003).

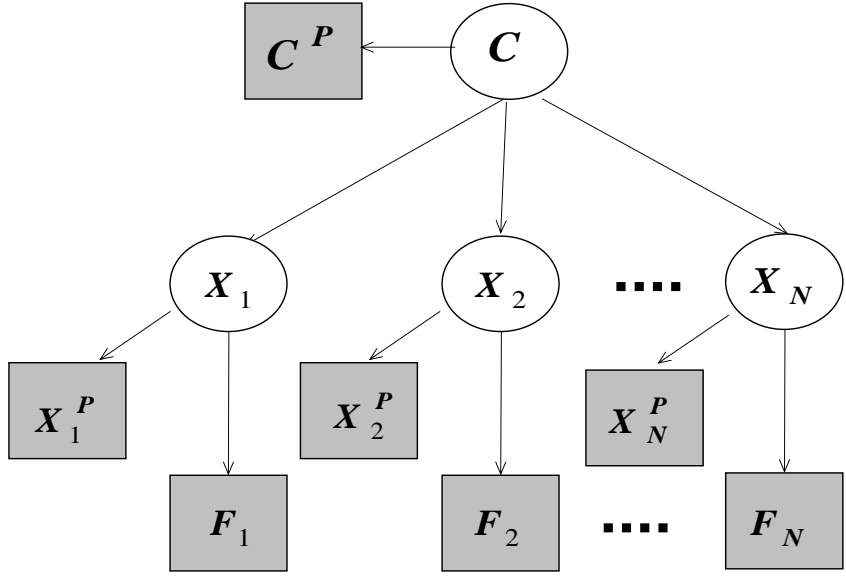

Figure 4: Probabilistic graphical representation of the two-frame recognition model. The latent variables C and {X} represent the image locations of the object and its parts in the current frame. The observed appearance of the parts in the current frame is represented by {F} which are image features. The observed image positions of the object and its parts at the previous frame are represented by $C^p$ and {$X^p$} respectively, while their measured velocities (derived from the optical flow between the frames) are represented by $V_c$ and {V} respectively.

The probabilistic framework of the two-frame spatiotemporal model is a natural extension of the initial static object detector and is defined as follows. At time frame $(t + \Delta t)$ we define a random variable C to represent the object center location in the image, and a set of random variables denoted by $\{X\} = \{X_i\}_{i=1}^{N}$ to represent the image locations of the N object parts. The observed appearance of the object parts in the image are represented by a set of random variables, which are image feature descriptors $\{F\} = \{F_i\}_{i=1}^{N}$. The interpreted image locations of the object and its parts in the previous frame t are represented by the random variables $C^p$ and $\{X^p\} = \{X_i^p\}_{i=1}^{N}$. The observed velocities of the object and parts are derived from the optical flow between frame t and $(t + \Delta t)$, and are represented by $V_c$ and $\{V\} = \{V_i\}_{i=1}^{N}$ respectively. The representation can be described by the graphical model shown in Figure 4. The full object and parts interpretation at frame $(t + \Delta t)$ is given by the joint probability as in (Eq. 1).

$$P(C, \{X\}, \{F\}, V_c, \{V\}, C^p, \{X^p\}) =$$
$$P(C) \cdot P(V_c) \cdot P(C^p/C, V_c) \cdot \prod_{i=1}^{N} P(X_i/C) \cdot P(V_i) \cdot P(X_i^p/X_i, V_i) \cdot P(F_i/X_i) \quad \text{(Eq. 1)}$$

**P($F_i/X_i$)**: We use a non-parametric representation for the conditional probability of the observed appearances of object parts $P(F_i/X_i)$. We use SIFT descriptors (Lowe, 2004) of image patches centered at object parts locations as appearance features. Given a set of part appearance features, the probability of a new appearance feature $F_i$ is obtained using a Gaussian kernel density estimation (KDE) over the L2 distances of $F_i$ from a subset of k nearest neighbors (k-NN) $\{Y\} = \{Y^j\}_{j=1}^k$ among the original set as shown in (Eq. 2). For efficiency we use approximate nearest neighbors search as in (Arya and Mount, 1993).

$$P(F_i/X_i) \approx \frac{1}{\sqrt{2\pi}hk} \cdot \sum_{j=1}^{k} \exp\left(-\frac{\|F_i - Y^j\|}{2h^2}\right)$$

(Eq. 2)

Using this non-parametric representation for the probability of the observed part appearances allow us to control the online adaption of the appearance model, by changing the set of known appearances at each time frame.

At the update phase of the two-frame scheme, concurrent appearances of successfully interpreted object parts at the previous frame are added to the current set of appearances, thus allowing a gradual adjustment of the appearance model via the k-NN approach. Furthermore, this approach provides a robust online adaption mechanism which can recover from possible erroneous interpretations, by memorizing previously observed appearances.

**P($X_i/C$)**: The structure of the object parts is represented as a geometric star-like model. The conditional probability of an object part given the object center $P(X_i/C)$, is modeled as a mixture of Gaussians: The first component is a Gaussian over spatial offsets between the object center and the part center in all images of the training set of the initial static object detector.

The second component is a Gaussian over similar spatial offsets which are being updated online during the update phase of the two-frame scheme while using recent object part interpretations.

The weight of the mixture components may be adjusted according to the interpretation confidence levels. However, in our experiments we used constant uniform mixing weights.

**$P(C^p/C, V_c), P(X_i^p/X_i, V_i)$**: Spatiotemporal consistency and motion constraints for the object and its parts between every two consecutive time frames, are represented via the conditional probabilities $P(C^p/C, V_c)$ and $P(X_i^p/X_i, V_i)$ respectively. For every two consecutive frames we calculate dense optical flow using the algorithm of (Black and Anandan, 1996). We then calculate the velocity of each interpreted object part and of the whole object at the previous frame t, as a weighted average of the optical flow at every pixel location within the part and object image regions respectively. These velocities imply Gaussian distributions for the location of the object and its parts at the current time frame $(t + \Delta t)$ given their interpreted locations at the previous frame t (Eq. 3).

$$P(C^p/C, V_c) \propto exp\left(-\frac{\|C - (C^p + (V_c \cdot \Delta t))\|}{2\sigma_c^2}\right)$$

$$P(X_i^p/X_i, V_i) \propto exp\left(-\frac{\|X_i - (X_i^p + (V_i \cdot \Delta t))\|}{2\sigma_i^2}\right)$$

(Eq. 3)

We assume uniform prior distributions for the object center among all image pixel locations $P(C)$.

## 2.3 Performance study

We demonstrate the performance of the adaptive parts detection model on two object categories: cars and airliners. The model is first trained on static images to learn object models at a particular pose. It is then applied to sets of video sequences containing instances of the class objects in various dynamic environments. At the beginning of each video sequence each object first appears at the same training pose. The objects then continuously change in appearance throughout the remaining of the video sequences.

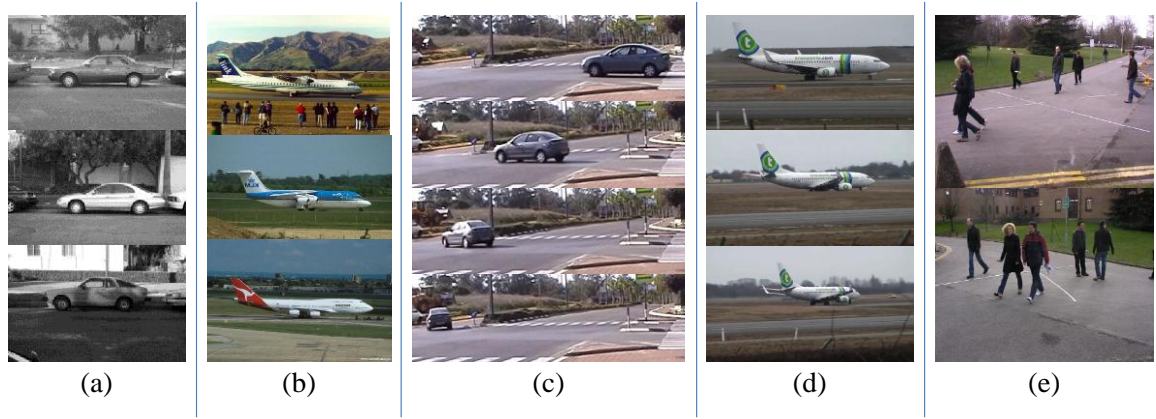

| (a) | (b) | (c) | (d) | (e) |

Figure 5: Image samples from the image and video datasets. (a+b) Training images used to learn an initial object model of a side view for the categories of cars and airliners respectively. (c+d) Test video sequences of cars and airliners undergoing changes in their viewing direction. (e) Test video sequences of non-class dynamic scenes for baseline comparison.

The cars video dataset consists of 3 video sequences with an average of 108 video frames per sequence. The cars appear initially from a side view and complete a turn ending with a rear-view by the end of the sequence.

The airliners video dataset consists of 4 video sequences with an average of 154 video frames per sequence. The airliners appear initially from a side view and change their pose during taxi or take-off.

A non-class background video test dataset was used for baseline detection performance evaluation emphasizing the robustness of the two-frame model on general video sequences which do not include instances of the known object. The dataset consists of 4 video sequences (extracted from PET2009 dataset) with an average of 110 video frames per sequence, and depicts people walking around on a road. Cars and airliners are not present in those scenes. Sample images from both training and test sets are shown in Figure 5.

### 2.3.1 Learning generic detectors for static images

For both categories a generic object and parts detection model was learned from a collection of side-view images containing the class objects (123 images of cars from Caltech101 dataset; 473 images of airliners) and a set of 467 non-class background images. Following the algorithm of (Ullman *et al.*, 2002; Vidal-Naquet and Ullman, 2003) 8 object parts were obtained for the cars category, 10 for the airliners.

We have tested the detection performance of our generic object detector on a validation set of the airliners category consisting of 100 images. Object detections in every image were compared to manually annotated ground-truth image locations.

### 2.3.2 Adaptive part detection

To test the two-frame adaptive parts detection model, we applied the model to every two consecutive frames in the test video sets. We have evaluated the detection performance of the object and all its constituent parts with respect to manually annotated image locations of the object and its parts at every video frame.

For comparison, the initial static object detector was also applied to each video frame, and evaluated for detection performance of the object and its parts.

Table 1 compares between the detection performance of the two-frame adaptive model and the initial object detector. Average precision rates (at the equal precision-recall point) and standard deviations were obtained for all test video sequences for each category. The results indicate an increase of more than 20% for cars and 30% for airliners, in the average precision rate of the two-frame adaptive model with respect to the single-image detector. The increase in performance was obtained for the full object as well as the individual parts. Sample parts detections from the test video sets are shown in Figure 6.

|  | Cars | | Airliners | |
| --- | --- | --- | --- | --- |
|  | Two-frame model | Initial Static model | Two-frame model | Initial Static model |
| Object | **77% ± 10** | 57% ± 12 | **99% ± 0** | 74% ± 19 |
| Part 1 | **57% ± 6** | 13% ± 17 | **78% ± 14** | 54% ± 22 |
| Part 2 | **78% ± 8** | 30% ± 13 | **60% ± 39** | 17% ± 18 |
| Part 3 | **25% ± 5** | 32% ± 13 | **71% ± 22** | 32% ± 18 |
| Part 4 | **76% ± 5** | 65% ± 5 | **78% ± 32** | 17% ± 10 |
| Part 5 | **77% ± 6** | 45% ± 13 | **85% ± 16** | 45% ± 30 |
| Part 6 | **66% ± 8** | 44% ± 4 | **95% ± 5** | 52% ± 38 |
| Part 7 | **53% ± 7** | 43% ± 3 | **59% ± 22** | 35% ± 27 |
| Part 8 | **75% ± 4** | 63% ± 4 | **69% ± 33** | 33% ± 24 |
| Part 9 | -- | -- | **81% ± 20** | 33% ± 16 |
| Part 10 | -- | -- | **68% ± 35** | 45% ± 16 |

Table 1: Adaptive parts detection. A comparison of object and parts detection performance between the two-frame object model and the initial static object model. Performance is evaluated by the average detection precision rate of the object and parts image positions, over the test video sequences, at the equal precision-recall point. The model consists of 8 parts for the cars category and 10 parts for airliners.

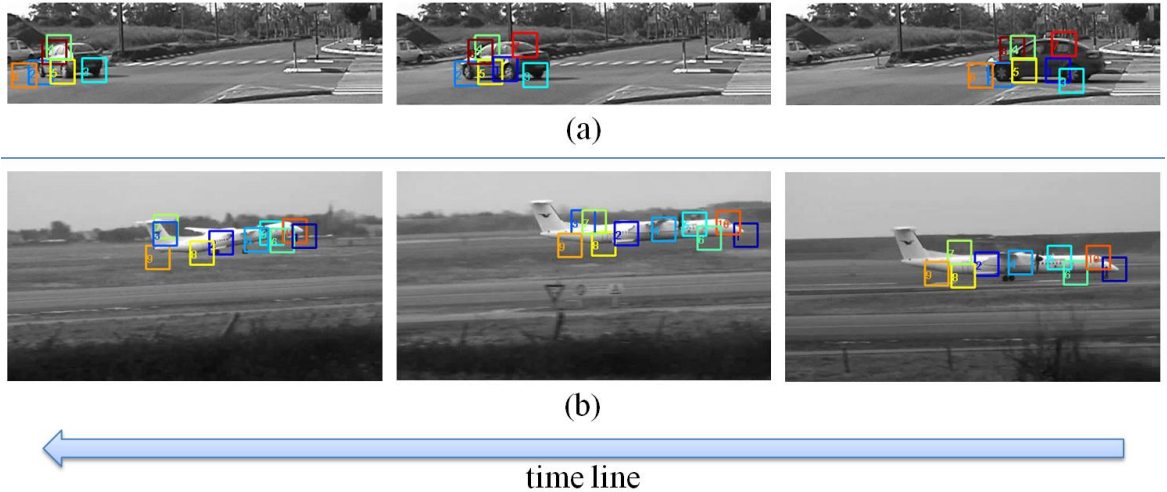

Figure 6: Parts detection examples of the two-frame model. (a) 'Car sequence 1' video at frames 1, 40, and 60. (b) 'Airlines sequence 3' video at frames 1, 50 and 100.

### 2.3.3 Robust adaption

Our online update scheme is gradual in the sense that the adapted model combines the old and current parts appearances and object geometry. The mixture is obtained by adding the appearance and displacement from the current model to the ANN structure. We compared this mixed adaptation with an alternative where the current-frame model (appearance and geometry of the detected object) completely replaces the previous model. Results are shown in Table 2 for two video sequences from the cars dataset.

The table shows the drop in detection precision rates for the object and its parts, relative to the robust adaption scheme, and even relative to the initial static detector for few parts.

|        | Cars sequence 1 | | | Cars sequence 2 | | |
|--------|:---:|:---:|:---:|:---:|:---:|:---:|
|        | Two-frames model | | Initial Static model | Two-frames model | | Initial Static model |
|        | Robust | Non-robust | | Robust | Non-robust | |
| Object | **70%** | **58%** | 44% | **72%** | **67%** | 53% |
| Part 1 | **50%** | **20%** | 20% | **55%** | **21%** | 2% |
| Part 2 | **79%** | **32%** | 12% | **66%** | **55%** | 41% |
| Part 3 | **19%** | **18%** | 25% | **24%** | **18%** | 20% |
| Part 4 | **77%** | **39%** | 65% | **82%** | **89%** | 60% |
| Part 5 | **74%** | **38%** | 43% | **85%** | **30%** | 45% |
| Part 6 | **62%** | **46%** | 40% | **78%** | **61%** | 43% |
| Part 7 | **45%** | **58%** | 42% | **51%** | **23%** | 40% |
| Part 8 | **74%** | **72%** | 65% | **81%** | **67%** | 57% |

Table 2: Gradual update scheme. A comparison of the detection precision rate at the equal precision-recall point, between the gradual update scheme of the two-frame model and an alternative update scheme. While the gradual update scheme combines previous and current parts appearance and geometry, the alternative update scheme completely replaces them after every successful detection. The detection performance is evaluated on 2 video sequences of the cars category and compared also to the initial static object model.

### 2.3.4 Detection prior is not enough

Temporal consistency of a dynamic visual input is an important source of information for the recognition process. Therefore, it may be argued that a static object model alone may suffice if we increase the chance for detecting the object after each successful detection.

To examine this possibility, we evaluated the detection of the static object detector on test video sets, while decreasing the initial detection threshold by a fixed rate after each frame when the object was successfully detected.

Table 3 shows the evaluation results for 2 video sets of the cars category. We use threshold decreasing rates of 1% and 2%, and compare the performance with a non-decreasing (0%) threshold. The initial threshold was obtained at equal precision-recall rates of the static object detector when applied to the training set.

As the threshold decreasing rate goes up, recall rates increase as well, but precision drops rapidly, and the overall performance is inferior to the two-frame model.

|  | Detection threshold decay rate | Precision | Recall |
|---|---|---|---|
| Cars sequence 1 | 0% | 100% | 12.9% |
|  | 1% | 56.3% | 20.5% |
|  | 2% | 42.9% | 43.2% |
| Cars sequence 2 | 0% | 100% | 26.7% |
|  | 1% | 95.5% | 36.2% |
|  | 2% | 50.4% | 53.5% |

Table 3: Is a detection prior enough for successful detection? Detection performance evaluation of the initial static object model when applied to 2 test video sequences of the cars category. The detection threshold is reduced by a fixed rate after every successful detection implying an increasing confidence level of finding the object at subsequent frames. Performance results are compared with the precision rate of the two-frame model at the equal precision-recall point when applied to these videos.

### 2.3.5 Learning new poses

The two-frame recognition scheme gradually learns new poses and appearances of the class object from the visual input dynamics in without external supervision. Each updated version of the initial static model captures the new appearance and structure of the detected object and thus represents a new pose, slightly different from the pose in the previous frame.

We demonstrate this learning capability of our two-frame recognition scheme by applying one of the updated models obtained from the cars category video sets to a new set of car images extracted from the internet. These images depict cars in a pose half way between a side view and a rear view. We have also applied the initial static model to this set for comparison.

|  | New pose model | Initial static model |
|---|---|---|
| Object | **66%** | 40% |
| Part 1 | **35%** | 2% |
| Part 2 | **69%** | 69% |
| Part 3 | **5%** | 25% |
| Part 4 | **80%** | 63% |
| Part 5 | **71%** | 59% |
| Part 6 | **37%** | 57% |
| Part 7 | **15%** | 6% |
| Part 8 | **79%** | 45% |

Table 4: Learning novel poses. Object detection precision rate at the equal precision-recall point of a learned new pose object model. The model was obtained from the two-frame model at frame 48 of the 'Cars sequence 1' video and applied to the 'Cars side-rear' image dataset. The detection performance is compared with the initial object model on the 'Cars side-rear' dataset.

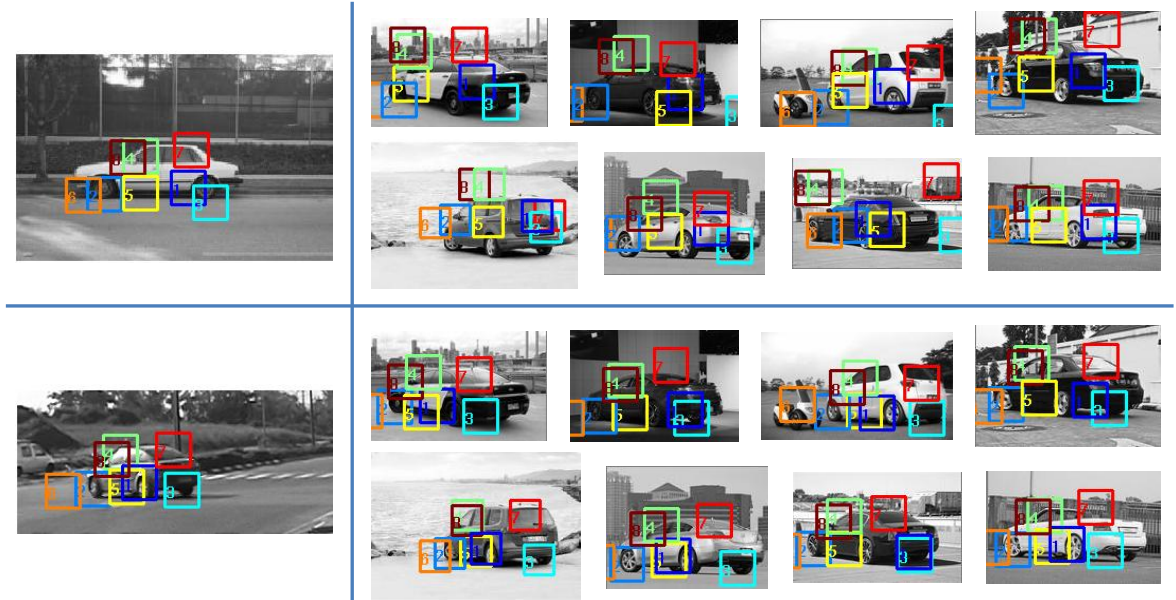

Figure 7: Parts detection examples of a new pose model on 'Cars side-rear' image dataset. The model was learned at frame 48 of the 'Cars sequence 1' video by the two-frame model. (a) Parts configuration of the initial object model. (c) Parts configuration as learned by the two-frame model at frame 48 of the 'Cars sequence 1' video. (b) Parts detection of the initial object model. (d) Parts detection of the new pose model.

Table 4 shows the detection precision rate of the object and its parts by the two models at the equal precision-recall point. Detection examples are shown in Figure 7.

## 2.4 Discussion

This work presents an approach to adaptive object and parts detection in dynamic environments. The dynamic changes are challenging, since both the appearance as well as the relative positions of visual features may change over time. Starting with an initial object model at some time $t_0$, we suggest an extended object model that continues to reliably detect the object and all of its parts at later times $t > t_0$, while gradually adapting to the changing appearance and structure.

We combine two sources of information in constructing the object model at time $(t + \Delta t)$: compatibility with the measured optical flow between time frame t and $(t + \Delta t)$, and similarity to the object model at time t. These sources of dynamic visual information are well studied in human vision and known as motion and spatiotemporal consistency. We suggest a simple new way of online updating the object model by an adaptive approximate nearest neighbors search and a statistical kernel density estimation.

The results show that the two-frame object model successfully extends the recognition capabilities of the initial object model. A comparison between the performance evaluation results of the two models demonstrates the superiority of the two-frame object model over the initial model for detecting the object and its parts in the test video sequences. This is true even when we use the initial static model while increasing its prior confidence level for detecting the object by a fixed rate after every successful detection.

We compared our gradual update scheme which combines previous and current parts appearance and geometry, with an alternative of completely replacing both the appearance and geometry of the detected object at each frame. The results show how our gradual combined update approach outperforms the detection performance of the replacement update alternative, which often 'drifts' away from the true object parts' model.

Finally, we demonstrate how the two-frame model adaption can also be used for learning with no external supervision, by enriching the initial model's recognition capabilities to a new set of viewing directions.

This approach can be further extended to automatically acquire full 3D understanding of the object and its parts from the dynamics of the visual input.

It will be interesting for future research, to base the entire scheme on dynamic input without an initial stage of learning a static model from a training image set, and to combine views across more than two successive frames.

# 3 'Mover' – Learning hands from a simple proto-concept

## 3.1 Overview

In chapter 2 we introduced an approach to adaptive object and parts detection in dynamic environments. We combined two sources of dynamic visual information, motion and spatiotemporal consistency, in constructing the object model and learning extended recognition capabilities with no external supervision.

In this chapter we explore the use of image motion in a most challenging recognition task – the task of learning to recognize human body parts, and in particular, hands. This task is of high computational difficulty, and was described by Mori *et al.* as "arguably the most difficult recognition problem in computer vision" (Mori *et al.*, 2004). Here, wwe suggest an unsupervised learning method for the automatic detection of hands in complex natural scenes. We refer to this unsupervised learning approach as internal supervision, since our learning process is applied to natural video sequences without external supervision.

Since hands are frequently engaged in motion, their motion could provide a useful cue for acquiring hand-concepts. We use image motion as an internal teaching signal, via the detection of 'mover' events. We define a 'mover' event as the event of a moving image region causing a stationary region to change after contact.

Psychological studies show that detecting hands (Yoshida and Smith, 2008), paying attention to what they are doing, (Aslin, 2009) and using them to make inferences and predictions (Gergely *et al.*, 2002; Saxe *et al.*, 2005; Sommerville *et al.*, 2005; Falck-Ytter *et al.*, 2006), are natural for humans and appear early in development. A large body of developmental studies has suggested that the human cognitive system is equipped through evolution with basic innate structures that facilitate the acquisition of meaningful concepts and categories (Piaget, 1952; Spelke and Kinzler, 2007; Carey, 2009; Meltzoff *et al.*, 2009; Tenenbaum *et al.*, 2011). These are likely not to be developed concepts, but simpler 'proto-concepts', which serve as anchor points and initial directions for the subsequent development of more mature concepts.

Infants are known to have mechanisms for detecting motion, separating moving regions from a stationary background, and tracking a moving region (Kremenitzer *et al.*, 1979; Kaufmann, 1987). However, our simulations show that general motion cues on their own are unlikely to provide a sufficiently specific cue for hand-learning (section 3.5). Infants are also sensitive, however, to specific types of motion: self-propelled, active (causing other objects to move), or passive (Michotte, 1963; Leslie, 1984; Luo and Baillargeon, 2005). These findings support our hypothesis of an innate domain-specific bias in the form of 'mover' events. 'Mover' detection is simple and primitive, based directly on image motion without requiring object detection or region segmentation.

The main idea behind the algorithm is fairly simple: Each video frame is partitioned into a grid of square cells. An object interaction is detected when an object enters a cell then exits, leaving the cell looking different than before. For example a hand may approach a cell containing a toy bear on the table, pick up the bear and move on. The cell is now changed, depicting the table without the bear (Figure 8). Instead of actually tracking all moving objects, we only follow the flow of individual pixels between cells. This proved to be a more robust approach for our purpose.

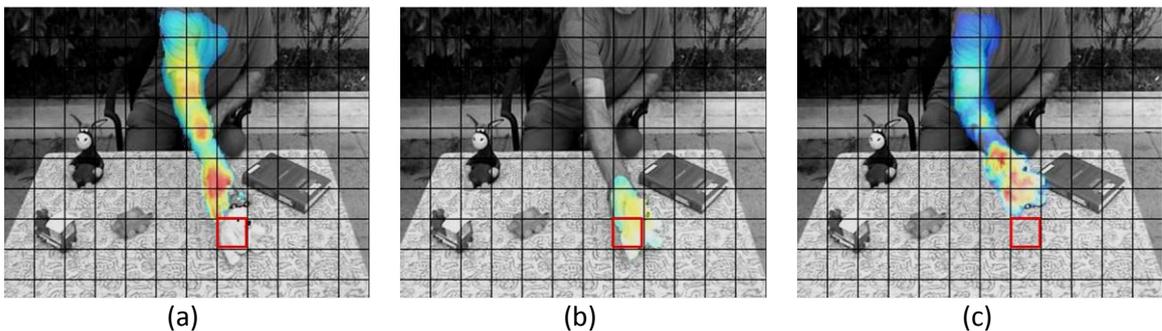

(a)          (b)          (c)

Figure 8: 'Mover' event. 'Mover' event detected in the red cell: Motion flows into the cell (a), stays briefly in the cell (b), and then leaves the cell, changing its appearance (c). Motion is shown in color, where warmer colors indicate faster motion.

## 3.2 Data

The data consists of 4 black and white videos sequences (see Figure 9), with a total of 22,545 frames (about 15 minutes). Each video frame is 360x288 pixels. Videos were taken with a static camera from the same viewpoint over the same background. An effort was made to minimize shadows and reflections, as these raise computational problems in motion and change detection that are not in the scope of this work.

Videos depicted one of three different individuals manipulating objects on a table. In addition to the motion of manipulating actions initiated by the actor, some objects were also invisibly propelled in order to imitate autonomous motion. The sources of motion were therefore agent-object interactions, ongoing intransitive movement of the actor, autonomous object movement and some background motion (i.e. breeze through leaves). The video sequences contained a total of 68 pickup actions, 60 put down actions and 67 events of autonomous object motion. They also contained 15 events where an autonomous moving object bumped into another object and moved it, effectively making it an agent.

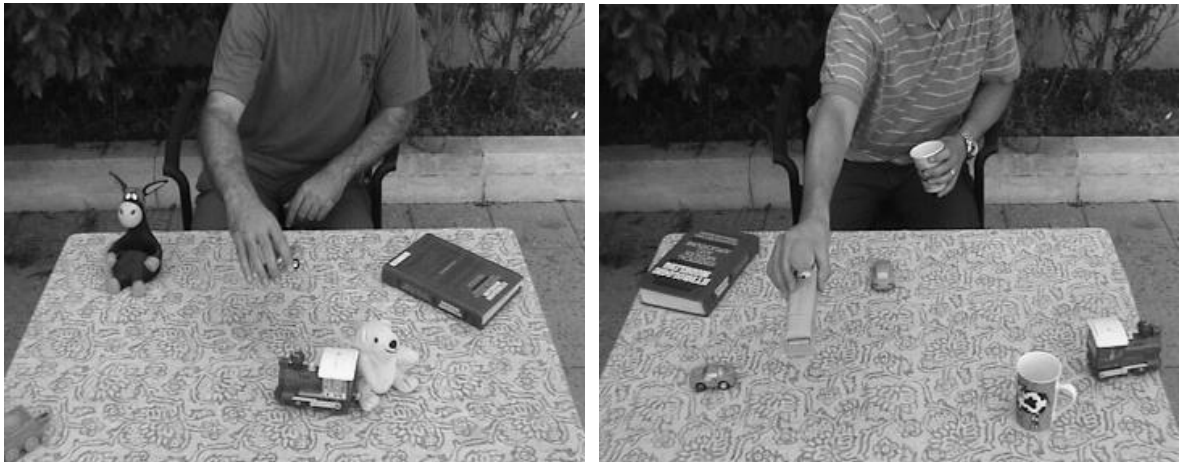

Figure 9: Sample frames from the video sequences used for 'mover' detection.

## 3.3  The algorithm

In section 3.1 we have described a simple and naïve method for detecting 'mover' events: detect incoming motion followed by outgoing motion, then check for changes. In practice, this task is more challenging: There are objects that span more than one cell, motions that go back and forth between neighboring cells (such as leaves in the breeze), slight changes in lighting, random camera noise, and even moving shadows. To overcome such difficulties, the detector needs to be somewhat more involved.

There are three typical types of motion that can cause false detection of 'mover' events:

1. Onset of autonomous motion – If an object spans more than one cell, some of these cells may exhibit incoming motion followed by outgoing motion, and will be changed when the object leaves them.
2. Ending of autonomous motion – Similarly, an object that spans more than one cell may exhibit incoming and outgoing motions when stopping, and change the cells it stops in.
3. Large body motion – Non rigid motion that spans multiple cells can cause random incoming motions, outgoing motions and changes in appearance.

We handle each of these cases separately:

1. When rigid objects begin moving, all of their parts start moving simultaneously. We require the onset of motion to precede the incoming motion into the cell.
2. After detecting the 'mover'-object interaction, we track the potential 'mover'. If it does not move we ignore it.
3. If the area of moving pixels within the cells of the detected 'mover' is too large, we ignore the 'mover'.

### 3.3.1 Grid of cells

We divide each frame into a grid of cells. Each cell is 30x30 pixels. The agent detector uses a state machine that works on each cell separately.

### 3.3.2 Background models (short term, long term)

For each cell we keep both a long term background and a short term background. The short term background is used to identify when a cell is changing (see below). In general it is updated every frame using an exponential moving average in order to account for noise and gradual lighting changes. However, when a cell differs significantly from the short term background it is designated as **changed**, and then the short term background is reset to the current appearance of the cell. The long term background is used for deciding whether a cell was changed by a potential agent. It stores the steady appearance of the cell, and is updated to reflect current cell appearance as long as the cell is **stable** (has not changed for some time, see below).

### 3.3.3 Detecting change

A cell is considered **changed** if at least 5 pixels differ from the short term background by at least 0.1 (on a brightness scale of 0-1). A cell is different from the long term background if the gradient magnitudes of at least 5 pixels differ from the background by at least 0.1. The long term comparison relies on gradients in order to ignore changes in lighting over time.

### 3.3.4 Moving pixels, incoming & outgoing motion

We use the optical flow algorithm by (Black and Anandan, 1996) to identify pixel motion between every two consecutive frames. If at least 5 pixels moved into a cell, we say that the cell had **incoming motion**. Similarly, if at least 5 pixels moved out of a cell, we say that the cell had **outgoing motion** (see Figure 10). A cell that had an **incoming motion** at least 3 frames ago (0.12 seconds) is considered **mobile** and remains **mobile** until it becomes **stable** again. If the **incoming motion** originates from **mobile** cells, we call it a **mobile incoming motion**. This helps us ignore autonomous motion that spans multiple cells as such motion will typically originate from **immobile** cells.

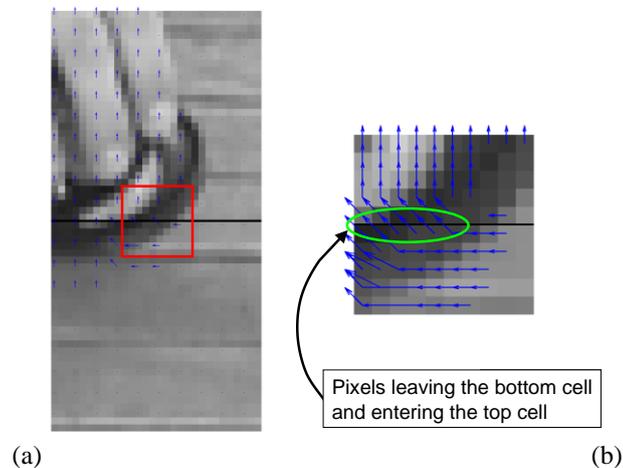

(a)                                                  (b)

Figure 10: Motion flow visualization. (a) The direction of the motion flow between two consecutive image frames is overlaid on the second image frame. The grid indicates the cells boundaries. (b) Zoom on two neighboring cells with cross motion flow. The upper cell has both incoming and outgoing motion while the lower cell has only outgoing motion.

### 3.3.5 Cell states: stable, incoming, outgoing, rejected

The state machine consists of the following states (see Figure 11 and Figure 12):

*Rejected* –        The cell may include some motion, but this is not a potential 'mover'. All cells start in this state.

*Stable* – The cell has not *changed* for some time. It does not appear to contain moving objects.

*Start changing* – The cell recently left the *stable* state. As the initial change might be due to shadows or reflections, this allows a small time window before the incoming motion. A **mobile** incoming motion will lead to the *incoming* state. An **immobile** incoming motion will lead to the *rejected* state. If there is no incoming motion in a few frames it will also lead to the *rejected* state.

*Incoming* – There has been an incoming motion of a potential 'mover'. An outgoing motion will lead to the *outgoing* state.

*Outgoing* – There has been an incoming motion followed by an outgoing motion. If the cell becomes *stable* soon, we may have detected a 'mover'. If too much time has passed since the cell left the *stable* state, change to *rejected* state as something is probably amiss. If too much time has passed since the last outgoing motion (and the cell is still not *stable*), the outgoing motion is probably not be the 'mover', return to *incoming* state.

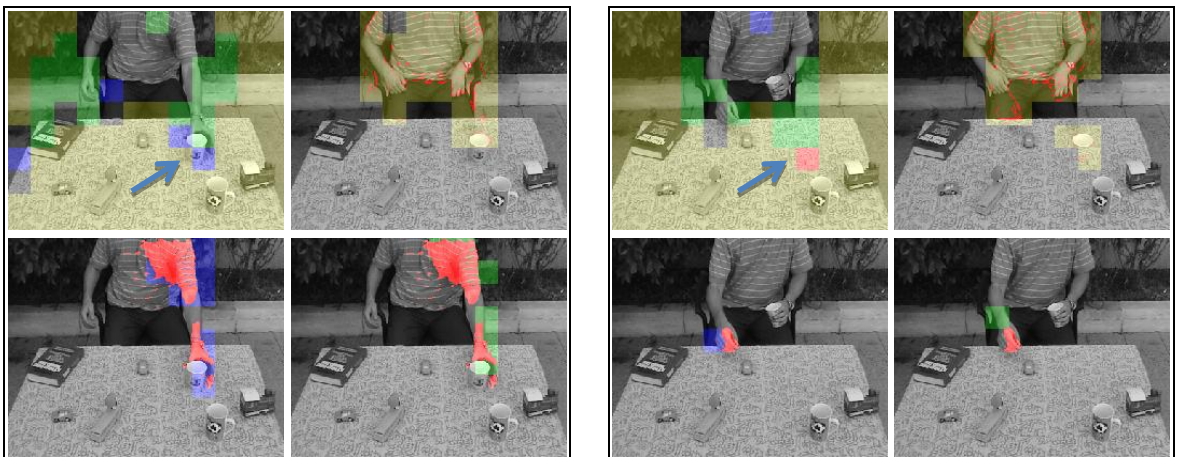

Figure 11: States of the 'mover' detector. Left: Hand approaches to picks up an object (marked with arrow). Right: 'mover' is detected several frames after the hand leaves the area. Legend: Top left: Cell states, yellow – stable, blue – incoming, green – outgoing, red – 'mover' detected, uncolored – rejected. Top right: yellow – cells that differ from background, red – pixels on edges that are different from background. Bottom: blue – cells with incoming motion, green – cells with outgoing motion, red – moving pixels.

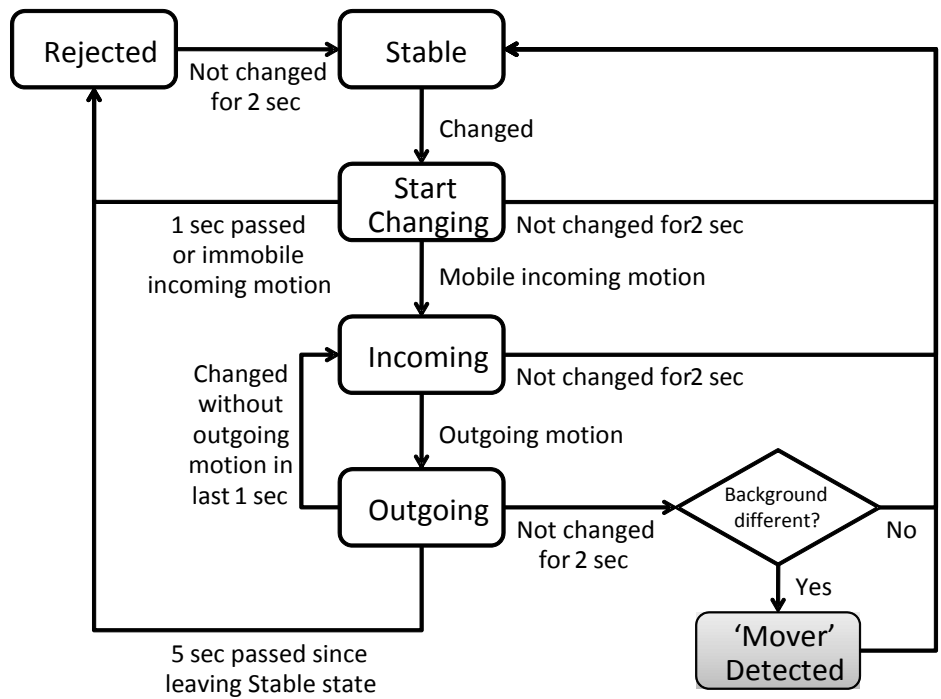

Figure 12: State machine for detecting 'mover' events.

### 3.3.6 Detecting 'mover'

When a cell in **outgoing** state becomes **stable**, we have hopefully seen an object entering the cell and then leaving it. If the cell is different from the long term background we detect a potential 'mover'.

The 'mover' detector may falsely detect a 'mover' due to large scale motion. In order to avoid such false alarms, we check for large scale motion in the vicinity of the cell, and ignore the detection if we find such motion. More precisely we ignore the detection if at least 50% of the pixels in the 3x3 cell area have optical flow magnitude of 0.2 or more.

### 3.3.7 Locking onto the 'mover'

Once a 'mover' is detected, we would like to find its center of mass. We search the 3x3 cell area for moving objects, and select the center of mass of the largest moving object. To find moving objects we consider all pixels with optical flow magnitude of at least 1, and perform a morphological close on them, using a 5x5 mask. This yields a mask of moving objects, of which we choose the largest one.

### 3.3.8 Tracking the 'mover'

The detected 'mover' events are used for training a hand detector. We track each detected 'mover' in order to draw more training examples. For the tracking we take SIFT descriptors from a 30x30 pixels around the object center, and search for them in the surrounding 90x90 region in the next frame. The SIFT descriptors in the first frame are taken at the intersection edges (Canny, 1986) and a 5x5 pixels grid. The SIFT descriptors in the second frame are taken at every pixel in the region of interest. Each descriptor in the first frame is matched to the nearest descriptor in the second frame, using L2 distance. Descriptors that have moved more than 30 pixels, or did not move at all are discarded. The rest of the descriptors are weighted by the magnitude of their optical flow and vote for the object center in the second frame, according to their offset from the object center in the first frame. Each 'mover' is tracked for up to 2 seconds. Tracking may terminate earlier if the 'mover' stops moving for 1 second. If the 'mover' does not move at all, it will be ignored. In case of multiple contemporary detections, intersecting tracks are merged.

## 3.4 Results

The 'mover' detector truly detected 116 object interactions out of 128 actions of pickup and put down, yielding a recall rate of 90.6%. It also falsely detected 124 other events, yielding a precision rate of 48.3%. Out of the 124 false detections, 44 were on autonomous moving objects (often on objects that indeed hit and moved other objects), 73 were on the body of the actor, 5 were on the background and 2 involved hand-object interaction but were not detected at the correct time.

Tracking increased the number of samples to 7,766 in 178 distinct tracks (see Figure 13). 93 of these tracks (52.2%) containing 5,183 samples (66.7%) begin with hands. Some of them lost the hand during tracking.

To evaluate the algorithm at different scales we have applied the same detection parameters and a fixed cell size of 30×30, to different scaled versions of an arbitrary video sequence from the *Movers* dataset. The different scales range from 50% up to 200% of the original video frame size. The 'mover' detection algorithm yields similar results at all scales as shown in Table 5.

| Scale | Detection precision of 'cell'-events | Detection performance of spatiotemporal events | |
|---|---|---|---|
| | | Precision | Recall |
| 50% | 79% (37/47) | 82% (28/34) | 78% (28/36) |
| 66% | 71% (54/76) | 71% (29/41) | 81% (29/36) |
| 100% | 77% (101/132) | 64% (34/53) | 94% (34/36) |
| 150% | 78% (120/154) | 63% (34/54) | 94% (34/36) |
| 200% | 78% (93/119) | 70% (30/43) | 83% (30/36) |

Table 5: Performance evaluation of 'mover' detection at different scales. The performance is evaluated at the cell level as well as at the spatiotemporal event level.

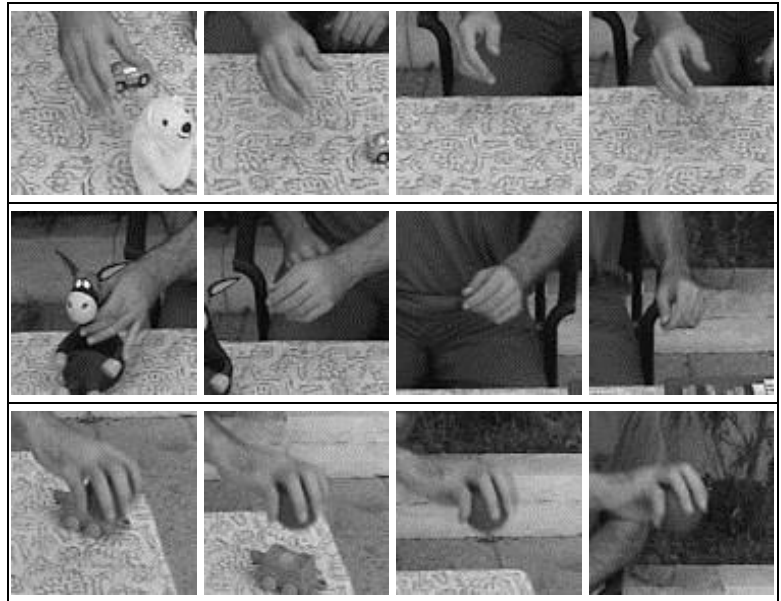

Figure 13: Additional hand appearances found by tracking. Each row is taken from one track. The leftmost image in each row was detected by a mover event. Only the central region in each extracted patch is used for tracking.

## 3.5 Trying alternative cues for learning hands

### 3.5.1 Overview

We use 'mover' events (active motion) as a cue for finding potential hand candidates. This choice conforms to results from developmental psychology (as discussed above), and also achieves good performance in our experiments. There are other possible cues that might be used to propose hand candidates in the image. We checked some prominent alternatives and report the results below.

We begin with several methods for capturing interesting image regions. The methods we compare are relatively simple, and may provide a more parsimonious cue than agency, since they are based on general cues for finding locations of interest, rather than 'mover' events that detect highly specific events.

1. Motion – Use any moving regions as hand candidates.
2. Space-time interest points – Use a 3D corner detector to find regions that change both in space and in time.
3. Image saliency – Extract prominent image regions as candidates.
4. General object prior – Use a generic object detector to suggest hand candidates from training movies.

We also consider the possibility that hand recognition follows (or is part of) learning to recognize an entire person. We assume the ability to identify images that contain a person. In real life this may be achieved by detecting a face, or by using other modalities such as voice, so this seems a reasonable assumption. Under this assumption, we check the hypothesis that hands may be learned as patches that are more representative of person images than of non-person images. We detect hand candidates using the method of informative fragments using person-images as 'class' and non-person images as 'non-class'.

As a last alternative, we check whether hands are an important body part for whole body person detection, and can therefore be learned as part of the process of learning to detect a person. We do this by training a state-of-the-art part based person detector, and checking whether successful detections consistently localize any of the learned parts with a hand. A detailed description of each method appears in the following sections.

We tested our 'mover'-based algorithm, the 4 different cues and the informative fragments method on a video sequence showing people walking, putting objects on a table and picking them up. The video is slightly over 10 minutes (around 15,000 frames). Each video frame is 360x288 pixels, and hands are typically around 30x40 pixels (Figure 14). In a realistic scenario general body motion is significantly more dominant than hand-objects manipulations. We therefore used a second video sequence showing people moving around with no special hand-object manipulations. This video is about 3 minutes long (around 4500 frames). Each video frame is 720x576 pixels, and hands are again typically around 30x40 pixels (Figure 15). We tested our 'mover'-based algorithm and the motion cue on a mixture of these two video sequences (taking an equal number of hand candidates from each video).

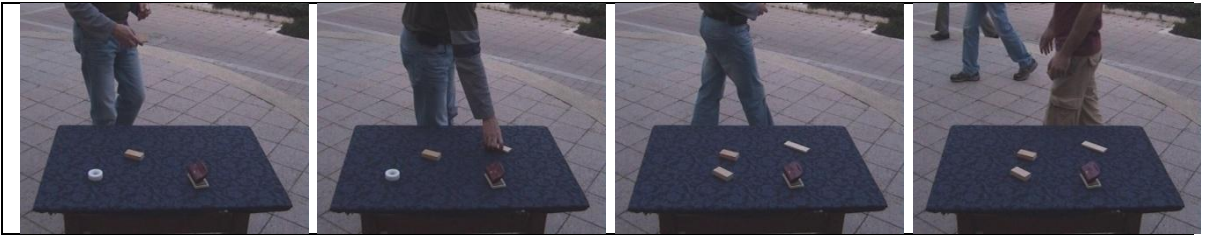

Figure 14: Alternative cues for learning hands. Samples from the training video sequence.

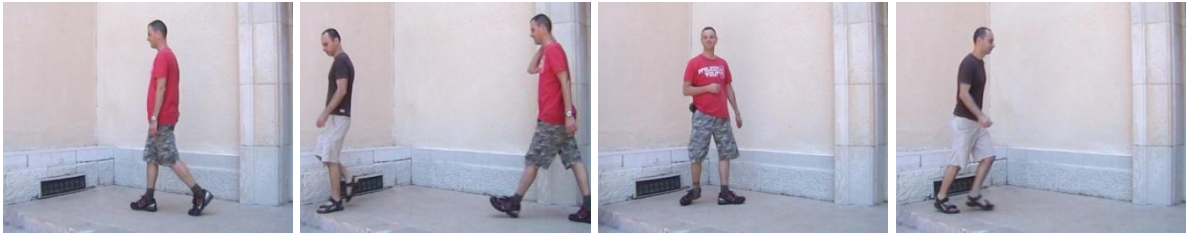

Figure 15: Samples from the 'Walk-about' video sequence, depicting general body motion.

We compared each hand candidate suggested by these methods to a manual ground truth annotation. For the general object prior, we considered a candidate to be true if it had a Jaccard score (ratio between intersection area and union area) of at least 0.25 with a 40x40 bounding box around the hand. For the other methods we considered a candidate to be true if the center-to-center distance between the candidate and the ground truth was less than 30 pixels.

Our 'mover'-based algorithm extracted 2883 hand candidates from the first testing video sequence, 63.6% of which were truly hands. No hand candidates were extracted by our algorithm from the second testing video sequence.

We compare this with the precision of the top 2500 hand candidates suggested by other methods. We also consider the option that some of these methods perform well only on the few best candidates. In order to check this we also report the precision on the top 100 candidates. The results are summarized in Table 6.

| | Precision for 100 best candidates | Precision for 2500 best candidates |
|---|---|---|
| **Motion** | 16.0% | 16.8% |
| **STIP (Laptev et al.)** | 27.0% | 38.1% |
| **Saliency (Walther & Koch)** | 0.0% | 0.48% |
| **Generic object detector ( Bogdan et al.)** | 0.0% | 0.2% |
| **Informative Fragments (Ullman et al.)** | 3.0% | 1.76% |
| **'Mover' + tracking (ours)** | | **64.7%** |

Table 6: Comparison of hand candidates' precision utilizing different methods. Hand candidates for general motion and 'mover' events are extracted from the two video sequences.

In order to check whether these results are sufficient for training a hand detector, we trained a supervised appearance based object detector (ANN-star) on the candidates extracted by each method. For training the detector, we used for class examples 2500 patches of 90x90 pixels taken around the top hand candidates suggested by each method. For non-class we used 2500 randomly chosen 90x90 patches adjacent to the class patches in the same original images.

We tested the performance of each detector on two different data sets:

1. *Manipulating hands* – 8 movies showing actors manipulating objects on a table (Figure 16).
2. *Freely-moving hands* – 8 movies showing actors moving their hands around freely in the air (Figure 17).

Each movie was between 40-80 seconds (1000-2000 frames); approximately 350x400 pixels (frame size changes slightly between movies to bring them into same scale). A detection is considered a hit if it falls within 30 pixels of the ground truth. The results are shown in Figure 18.

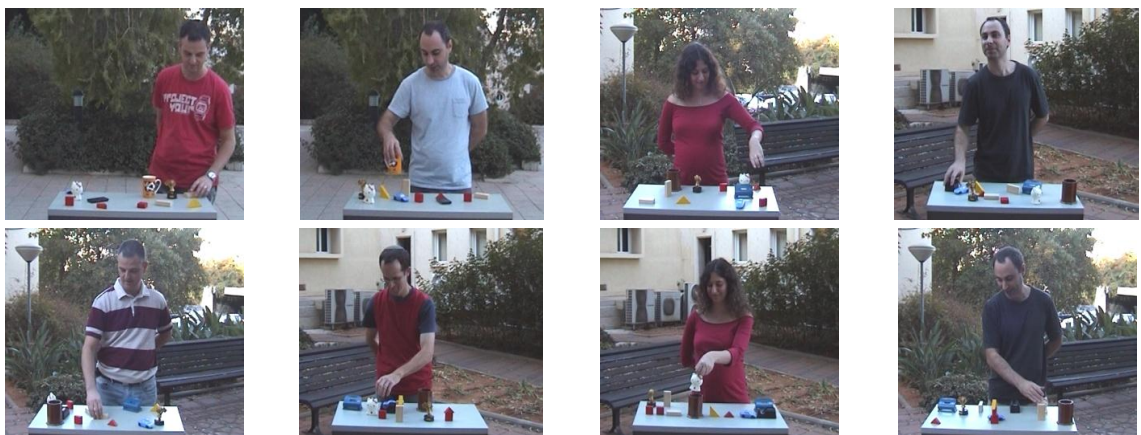

Figure 16: Example frames from the 8 *Manipulating hands* video sequences.

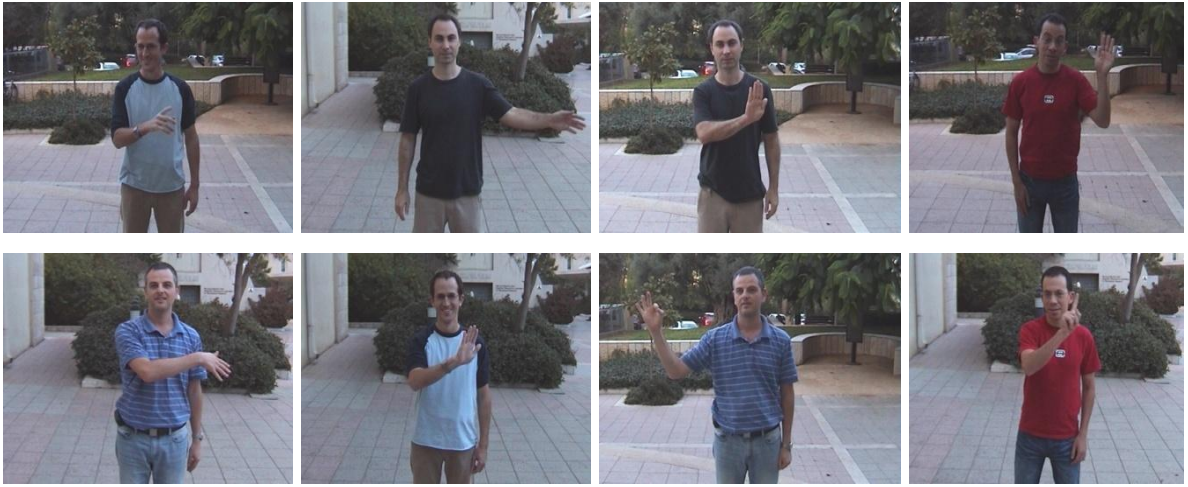

Figure 17: Example frames from the 8 *Freely-moving hands* video sequences.

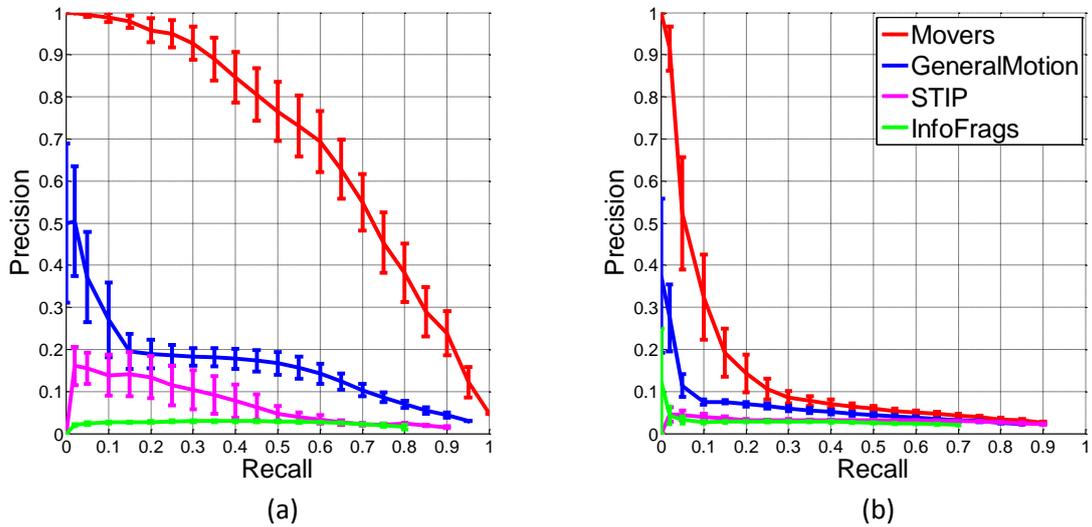

Figure 18: Precision-recall curves of hand detectors trained on hand candidates found by the different alternative cues. Performance is averaged over all video sequences of (a) the *Manipulating Hands* dataset, and (b) the *Freely-moving Hands* dataset. Red: training by 'mover' events, blue: training by general motion, magenta: training by spatiotemporal interest points, green: training by information maximization.

**Experiments**

Following is a detailed description of the studies summarized in the section 3.5.1 and Table 6.

### 3.5.2 Motion

Psychophysical studies show that young infants pay a lot of attention to motion. In fact, they attribute such significance to motion, that they would expect object boundaries to be defined by common motion more than by common form, color or texture (Spelke, 1990). Thus, we can strongly believe that infants use motion as a major source of information for learning hands.

To use the motion cue, we calculated the optical flow (Black 1996) for every video frame. For each frame we marked the point of maximum optical flow magnitude. The points with the highest optical flow magnitude between all frames were chosen as hand candidates (see examples in Figure 19 ).

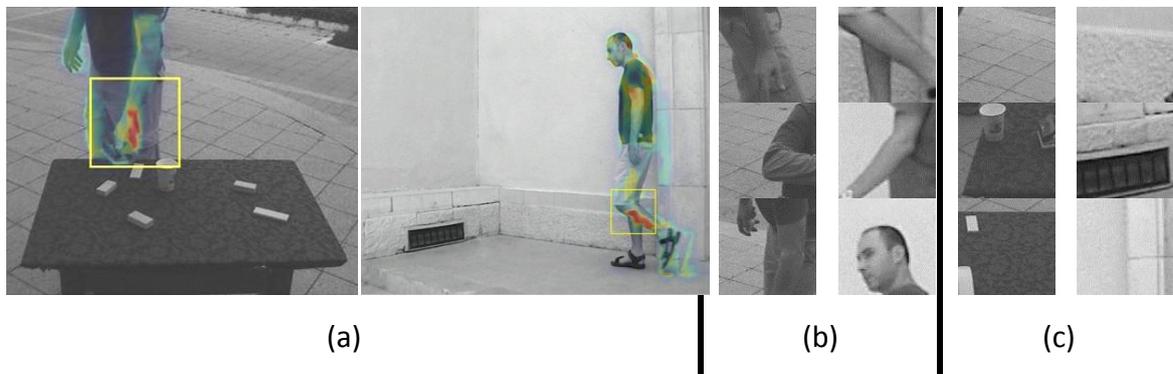

(a)  (b)  (c)

Figure 19 : Settings for the "high motion" training phase: (a) sample frames from the training video (optical flow overlaid) and extracted image patch around highest motion; (b) positive patches extracted around "highest motion" image locations; (c) negative patches extracted at random image locations.

### 3.5.3 Space-time interest points

Space time interest points extend the Harris corner detector to find points in video with large variation in both space and time (Laptev, 2005). A detected point will typically be located at the tip of a moving object at the peak of its motion. It seems plausible that the visual system, which is sensitive both to edges and motion, will also tend to focus on these points. Space-time interest points are likely to occur on fast moving protruding limbs, such as hands. We would like to check whether this may provide a strong enough cue for detection of hand candidates.

In order to detect candidates in the video sequences, we used the code provided by Laptev. This code detects space time interest points and assigns a score to each point. We selected the top scoring interest points as hand candidates (Figure 20).

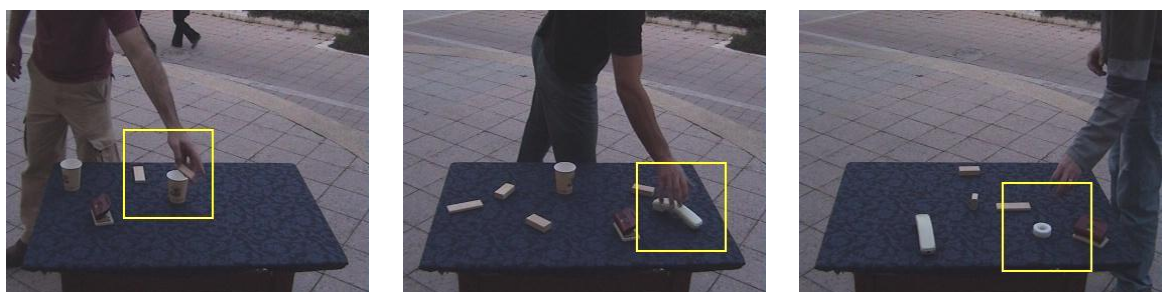

Figure 20: Examples of regions around space time interest points.

### 3.5.4 Saliency

Saliency is the state or quality by which an item stands out relative to its surroundings. Psychophysical and physiological evidence indicates that the visual system of primates and humans has evolved a specialized processing focus moving across the visual scene. This saliency detection capability is considered to be a key mechanism for selecting visual attention, which facilitates learning by enabling the focus of the limited perceptual and cognitive resources on the most pertinent subset of the available sensory data (Koch and Ullman, 1985).

Computational approaches suggest that a set of elementary features such as orientation, color or motion, is computed across the visual field and is represented in a set of cortical topographic maps. Locations in the visual space that differ from their surroundings with respect to an elementary feature are singled out in the corresponding map. Combining these maps into a saliency map, encode the relative conspicuity of the visual scene (Koch and Ullman, 1985).

We applied a state-of-the-art saliency detector (Walther and Koch, 2006) to each frame of the training video clips and extracted the 10 most salient image locations and their saliency score. We chose the top scoring salient regions as hand candidates. Figure 21 shows two training images and their salient locations.

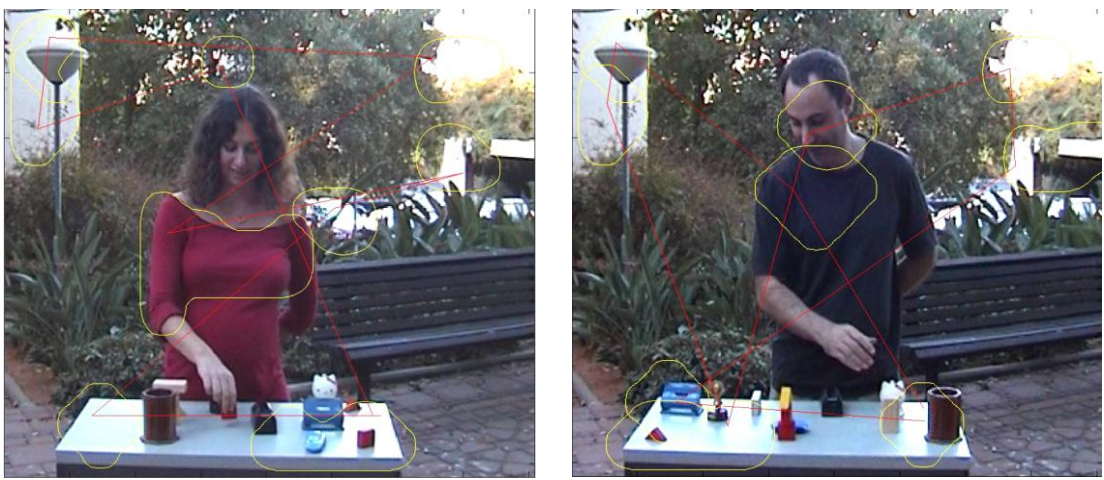

Figure 21: Samples images with detected salient regions from the video sequences. Salient regions are marked in yellow. Saliency sequential shifts are marked as red lines between extreme salient points.

### 3.5.5 Generic object detector

Ferrari et al. (Alexe *et al.*, 2010) propose a general measure of how likely it is for an image window to contain an object. This measure uses a Bayesian model to combine four different cues:

1. Multi-scale saliency – Using spectral residuals (Hou and Zhang, 2007).
2. Color contrast – The chi square distance between the LAB histograms of the window and its surrounding.
3. Edge density – Density of edges along the inner perimeter of the window.
4. Super-pixel straddling – Total area of super-pixel regions that cross the window boundary, relative to the window size.

According to the authors, their 'objectness' measure outperforms traditional saliency measures on the PASCAL VOC2007 dataset (Everingham *et al.*, 2007). They suggest this measure "can be useful also in other applications, such as to help learning object classes in a weakly supervised scenario, where object locations are unknown".

We ran the code by Ferrari et al. on our video sequence, and used the top scoring windows as hand candidates. Figure 22 shows some candidates suggested by the 'objectness' measure.

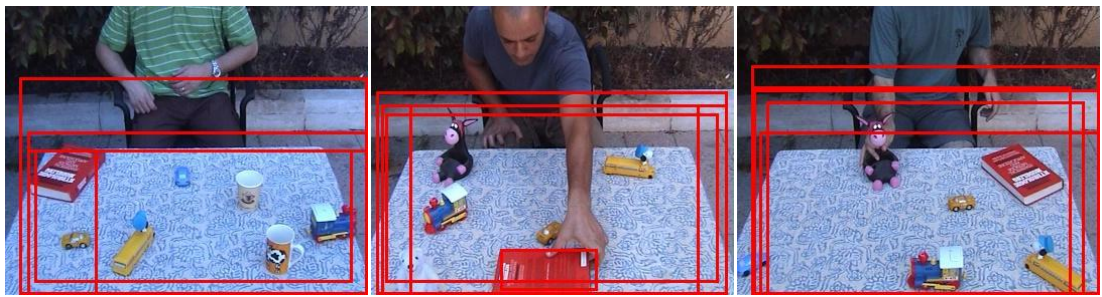

Figure 22: Sample images with detected generic objects from the video sequences. The red rectangles show the 5 windows with the highest 'objectness' score in each frame.

### 3.5.6 Informative Fragments

A general approach to the representation of object classes and to the task of visual classification is the fragment-based representation, which can also be applied to the class of hands. In this approach, images of objects within a class are represented in terms of class-specific fragments. These fragments provide common building blocks that can be used, in different combinations, to represent a large variety of different images of objects within the class (Ullman *et al.*, 2002). The fragments may be selected from a training set of images based on a criterion of maximizing the mutual information of the fragments and the class they represent (Vidal-Naquet and Ullman, 2003).

Unlike the previous method, we now need to compare class and non-class images. For class images we used the same video sequence as before, but restricted ourselves to the region containing the actor using background subtraction (we compared each frame to a basic background image that did not contain an actor). For non-class we used 300 images from the PASCAL VOC2007 challenge dataset (Everingham *et al.*, 2007). Figure 23 shows some examples of the training data. A random set of 166,979 image fragments was extracted from the training class images. We chose hand candidates from these fragments using the Max-Min greedy search algorithm that maximizes the additional information added by each selected fragment (Figure 24).

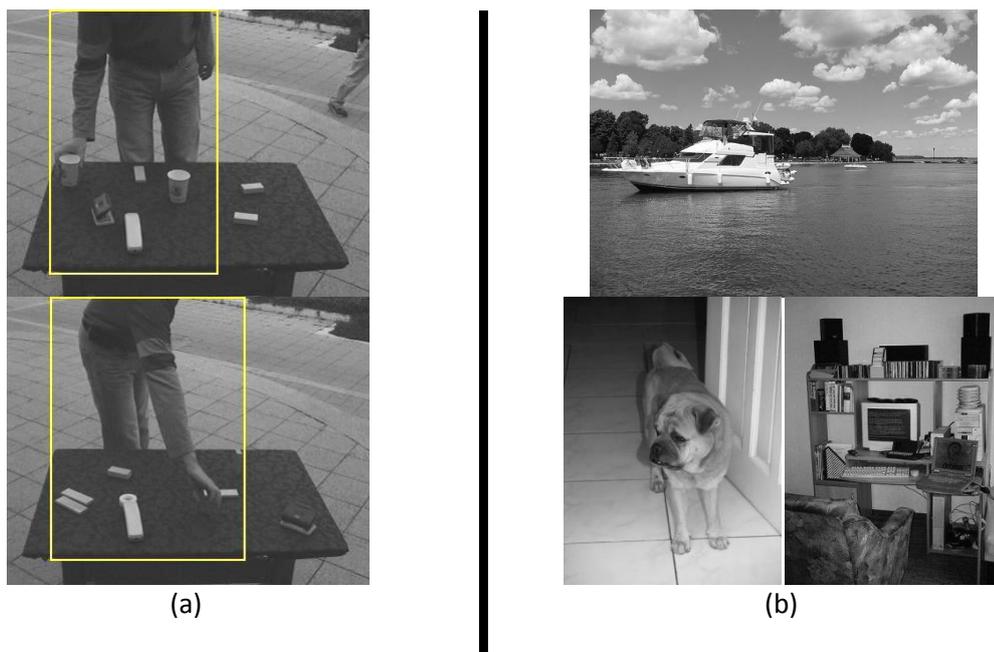

(a)  (b)

Figure 23: Sample training images for selecting person informative image fragments with the Max-Min search algorithm. (a) Person, (b) Non-person.

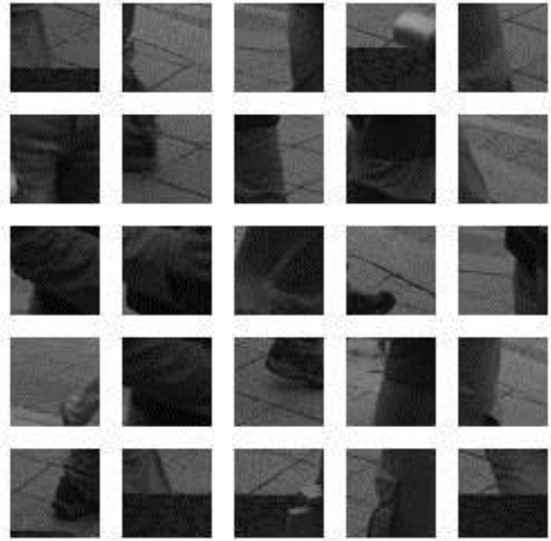

Figure 24: Resulting 25 most informative image fragments for the class object 'Person' (head excluded).

### 3.5.7 Part-based object recognition

As parts of the human body, hands may be learned by utilizing full body representation models. These models may automatically learn the appearance and geometrical configuration of constituent parts of a deformable object class, given the location and size of the object in a set of training images (Epshtein and Ullman, 2005; Felzenszwalb *et al.*, 2010).

For this experiment we trained a state-of-the-art object detector (Felzenszwalb *et al.*, 2010) with 8 unlabeled constituent parts. We used the *Manipulating hands* video (on which the other methods were tested) for class examples. We used background subtraction (as we did for the informative fragments) to create bounding box labels around the person. For non-class we used 1439 "non-person" images from the PASCAL VOC2007 dataset (Everingham *et al.*, 2007). We trained two separate models for right hand and left hand, using 1435 and 1432 class images respectively (the non-manipulating hand was hidden behind the back of the human performer). Figure 25 shows examples of the training data. The resulting upper-body object detection models are visualized in Figure 26.

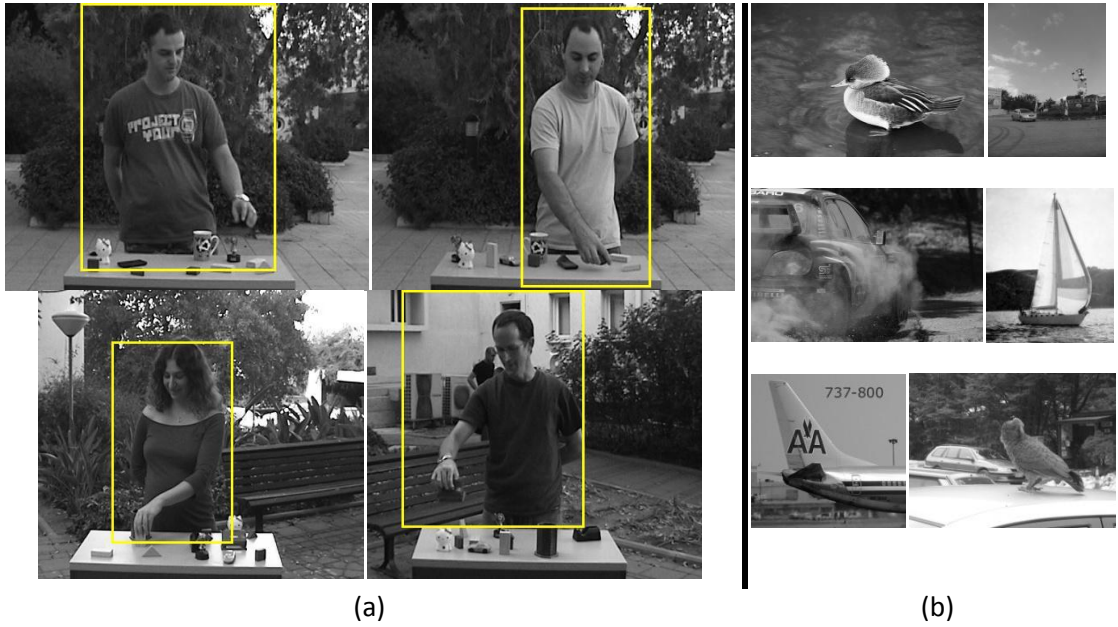

(a) (b)

Figure 25: Training data for the part-based object detection models: (a) positive image examples and annotated bounding-boxes for upper-body models with left and right hand manipulations; (b) negative examples of non-person images.

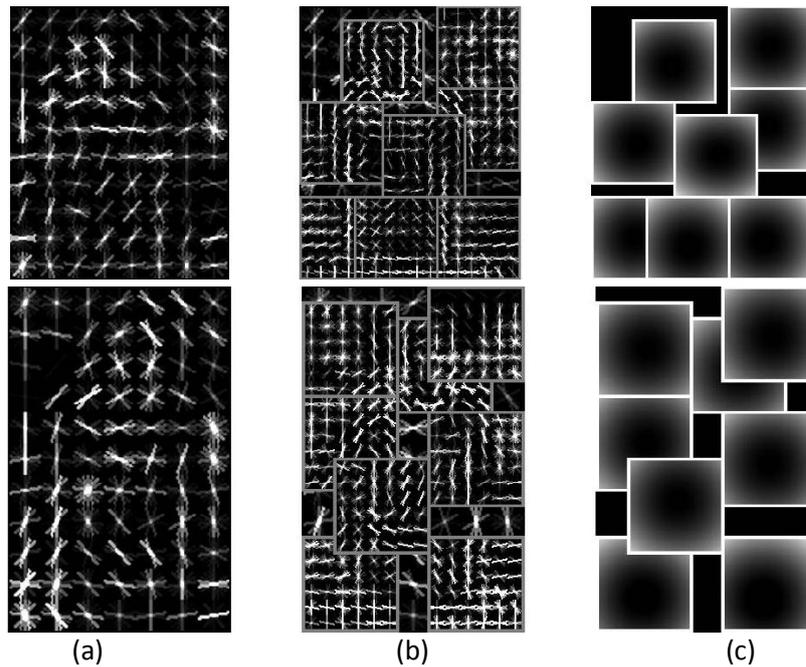

(a) (b) (c)

Figure 26: Visualization of the part-based upper-body models trained on the *Manipulating hand* dataset: (Upper row) upper body with right-hand model; (Lower row) upper body with left-hand model. Both models visualization consists of: (a) upper-body root filter; (b) body-parts appearance filters; (c) body-parts spatial configurations.

Next, we applied both right-hand and left-hand detectors to the training images of each model, and measured the precision and recall rates of detecting true hands within a predefined distance radius (30 pixels) from the center of each detected body part. The resulting average precision of both "left-" and "right-hand" upper-body models for the best "hand" part of each model was 65.38% (at an average recall rate of 1.19%). If we ignore the learned part appearance and use only the spatial configuration, we get an average precision of 35.8% (at an average recall rate of 14.2%). Thus, over half of the detected hands are found based on their spatial location with respect to the upper-body position, regardless of their appearance.

These results on the training images outperform the hand candidates' precision of the other methods (see Table 6). However, applying both detectors to the test video datasets, yield poor hand detection rates on both 'manipulating' and 'freely-moving' hands (Figure 27). These results indicate the over-fitting of the learned models to the training data, while not being able to generalize recognition capabilities.

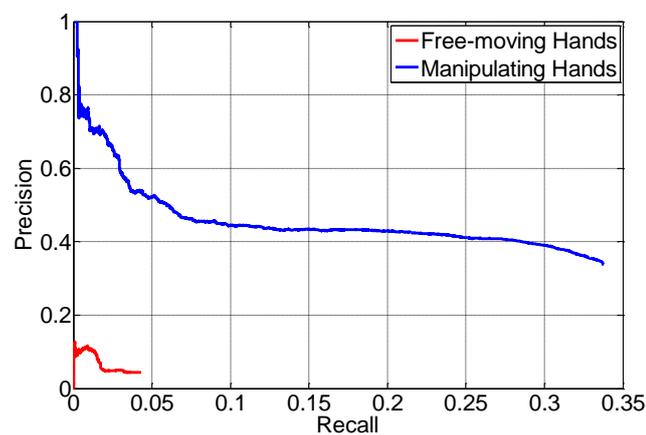

Figure 27: Precision-recall of hand detection utilizing Felzenszwalb's upper-body models on our two testing video sets.

## 3.6 Summary

Based on developmental research we have introduced in a model for learning hands, the detection of active motion, which we call a 'mover' event, defined as the event of a moving image region causing a stationary region to move or change after contact. 'Mover' detection is simple and primitive in the sense that it is based directly on image motion without requiring object detection or region segmentation. Because hands are highly represented in 'mover'-tagged regions (about 65% in our experiments), following this learning phase, hands can be detected with high precision, but with low recall rate (97% at 2% recall in our experiments). This is due to the fact that detection is limited to specific hand configurations extracted as 'mover' events, typically engaged in object grasping.

Our testing showed that current computational methods for general object detection (Ullman *et al.*, 2002; Walther and Koch, 2006; Felzenszwalb *et al.*, 2008; Alexe *et al.*, 2010) applied to large training data do not result by themselves in automatically learning about hands. In addition, our simulations showed that general motion cues on their own are unlikely to provide a sufficiently specific cue for hand-learning. Thus, we suggest that the incorporation of a plausible innate or early acquired bias, based on cognitive and perceptual findings, leads to the automatic acquisition of increasingly complex concepts and capabilities, such as learning to detect hands, which do not emerge without domain-specific biases.

In the next chapter we will introduce contextual information as an important cue for extending the recognition capabilities. We will show how the co-training of two complementary cues, context and appearance, is used for internal supervision and leads to continuous improvement of the initial ('mover'-based) detection performance in an unsupervised manner.

# 4 Co-training of complementary detectors to boost hand learning

## 4.1 Overview

In chapter 3 we introduced in a model the detection of active motion, which we call a 'mover' event. Since hands are highly represented in 'mover'-tagged regions, following this learning phase, hands can be classified with high precision, but at a low recall rate. This is due to the fact that 'mover'-based detection is limited to specific hand configurations typically engaged in object grasp and release.

In this chapter we show that the co-training of two complementary detectors using different cues leads to continuous improvement of the initial detection performance. In particular, we show that appearance-based object detection and context-based part detection can guide each other to boost recognition performance. Starting with limited hand configurations learned from 'mover' events, we introduce contextual information of neighboring body parts such as faces (Appendix A - The chains model for detecting parts by their context). The context-based detection successfully recognizes hands with novel appearances, provided that the pose is already known ('pose' here is the configuration of context features, on the shoulders, arms, etc.). The newly learned appearances lead in turn to the learning of additional poses. In this manner, context and appearance facilitate each other in a co-training cycle. This approach is also supported by developmental evidence indicating that infants associate hands with other body parts at around 6-9 months (Slaughter and Heron-Delaney, 2010), and possibly earlier with faces (Quinn and Eimas, 1996; Slaughter and Neary, 2011).

Detection performance improves rapidly in subsequent iterations of the learning algorithm, based on tracking and body context (as demonstrated in section 4.4). Detected hands are tracked over a short period of time, while additional hand image examples are extracted along the tracking trajectory. Developmental research supports the use of tracking in the learning process, as young infants are known to have such mechanisms as detecting motion, separating moving regions from a stationary background, and tracking a moving region (Kremenitzer *et al.*, 1979; Kaufmann, 1987; Spelke, 1990).

## 4.2 Data

We use two sets of unlabeled video sequences, one (denoted as *Movers* dataset) for learning an initial hand detector from 'mover' events, and another set (denoted as *Freely-moving hands*) for learning a combined appearance-context hand detector.

The *Movers* dataset consists of 4 black and white videos sequences, with a total of 22,545 frames (about 15 minutes). Each video frame is 360x288 pixels. Videos were taken with a static camera from the same viewpoint over the same background. The videos depicted one of three different individuals manipulating objects on a table (Figure 28).

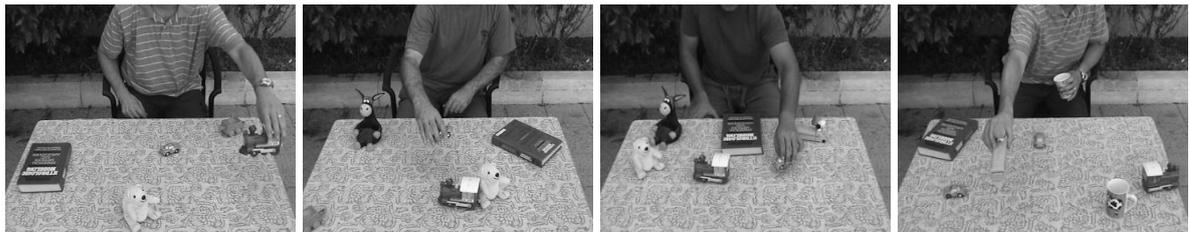

Figure 28: Sample frames from the *Movers* video dataset.

The *Freely-moving hands* dataset consists of 8 outdoor video scenes taken with a static video camera at a fixed exposure setting. The video sequences, about 60 seconds each, were converted to gray level. Each video frame is at a 436x336 pixels resolution. The video scenes depict one of four different human actors facing the camera in roughly two different backgrounds (camera position was slightly altered between shots). The performers were asked to move their hands around (one hand at a time) while changing hand gestures among roughly six different gestures: feast, thumbs up, forefinger extended up, V-sign, OK-sign, and stop-sign. The performers were asked also to presume free natural hand poses occasionally during the whole maneuver which lasts about 30 seconds per each hand (stationary hand is resting along the body side). We denote this video dataset as the *Freely-moving hands* dataset (Figure 29).

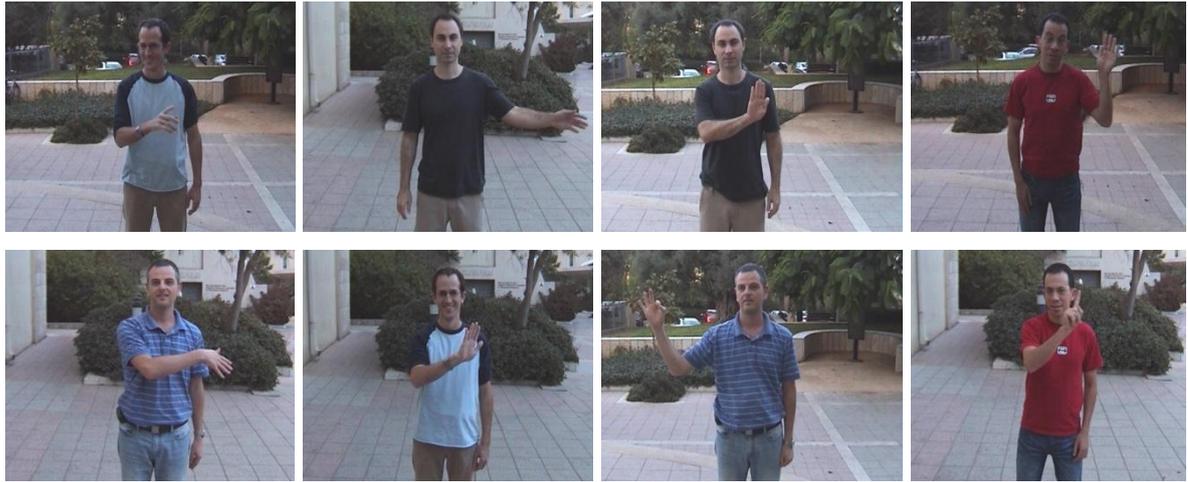

Figure 29: Sample frames from the *Freely-moving hands* video dataset.

## 4.3 The algorithm

### 4.3.1 General

The initial 'mover'-based hand detector automatically improves by combining local appearance and body context. All learning is completely unsupervised. The algorithm is applied in iterations, each goes once over the full training data. Each iteration consists of the following steps:

(i) Detections by the detector from the previous iteration are tracked for up to 2 seconds, and provide new training examples. (The first iteration uses the mover-based detector.)

(ii) An appearance-based classifier and a context-based classifier are trained.

(iii) Detection scores are combined to produce the improved detector for the next iteration.

Final detectors are evaluated on new movies which are not used during learning (Figure 30).

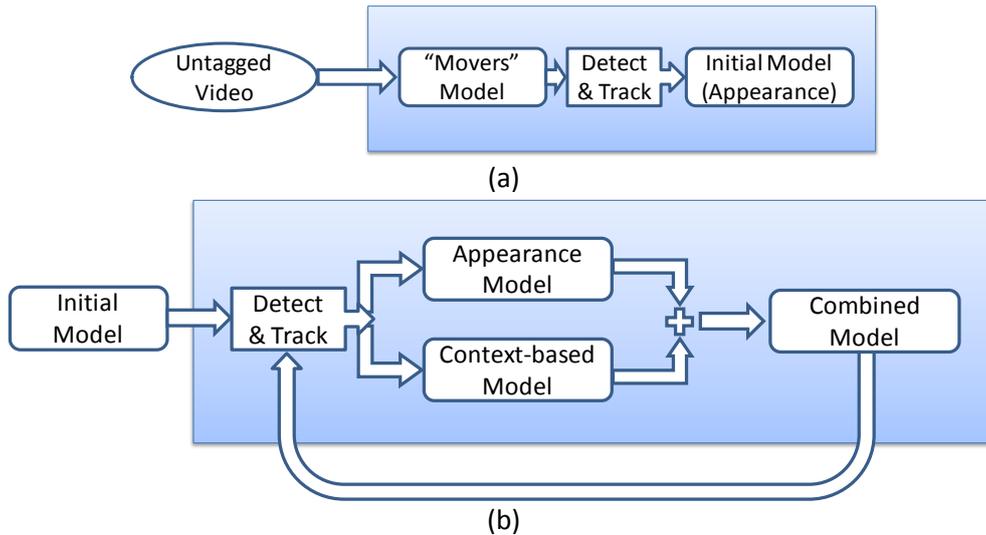

Figure 30: Schematic flow our hands learning process. (a) Learning an initial appearance model from detected 'mover' events. (b) The appearance-context learning cycle.

We learn a geometric star model for the appearance classifier, and a context part-based geometric model for the context-based classifier (Karlinsky *et al.*, 2010). Both models utilize dense SIFT features along object boundaries to represent the image data.

For tracking we use a simple object tracker which tracks an object (represented by dense SIFT features) between every two consecutive video frames while utilizing the optical flow at the vicinity of the object (Black and Anandan, 1996).

### 4.3.2   Initial learning stage

Applying the 'mover'-based hand detector to a training video sequence, we first extract 90x90 pixels image patches centered at the location of each 'mover' event. Utilizing optical flow and object tracking techniques, we extract additional image patches along the trajectories (2 seconds) of the detected 'mover' events. We use these image patches as positive hand examples (Figure 31a), and extract also negative background image patches (Figure 31b), for learning an initial hands appearance model (Figure 30a). We utilize the appearance-based object detector (ANN-star) presented in Appendix A - The chains model for detecting parts by their context (Karlinsky *et al.*, 2010). This hand appearance model is a weak general hand detector, but detects hands at grasping and reaching poses (which are common to 'mover' events) with high confidence level.

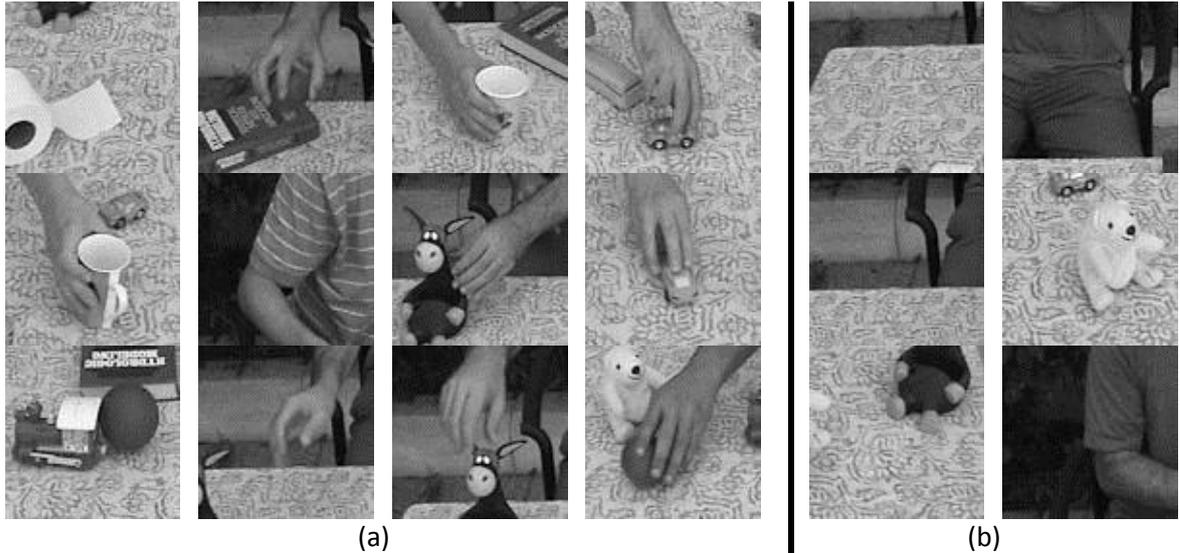

(a)                         (b)

Figure 31: Image patch examples used to train the initial appearance-based hand detector. (a) Positive examples; (b) Negative background examples.

### 4.3.3 Appearance learning stage

Given a weak hand detector, we utilize its highest detections when applied to our training video sets to re-learn an improved appearance-based hand detector. This learning stage is conducted in an internally supervised learning framework, in which a given set of positive and negative image examples is used to train a new appearance model (ANN-star) for hand detection. The positive image examples are 90x90 pixels image patches extracted from training video frames around high score detections of the weak hand detector. Negative image examples are background 90x90 pixels image patches, extracted from the same video frames away from the detections of the weak hand detector (Figure 32). Both positive and negative examples are noisy, since the detections of the weak hand detector include both true and false positive examples (i.e. we may have non-hands positive image patches and negative image patches that consist of hand instances). No external labeling is used during training in this learning stage.

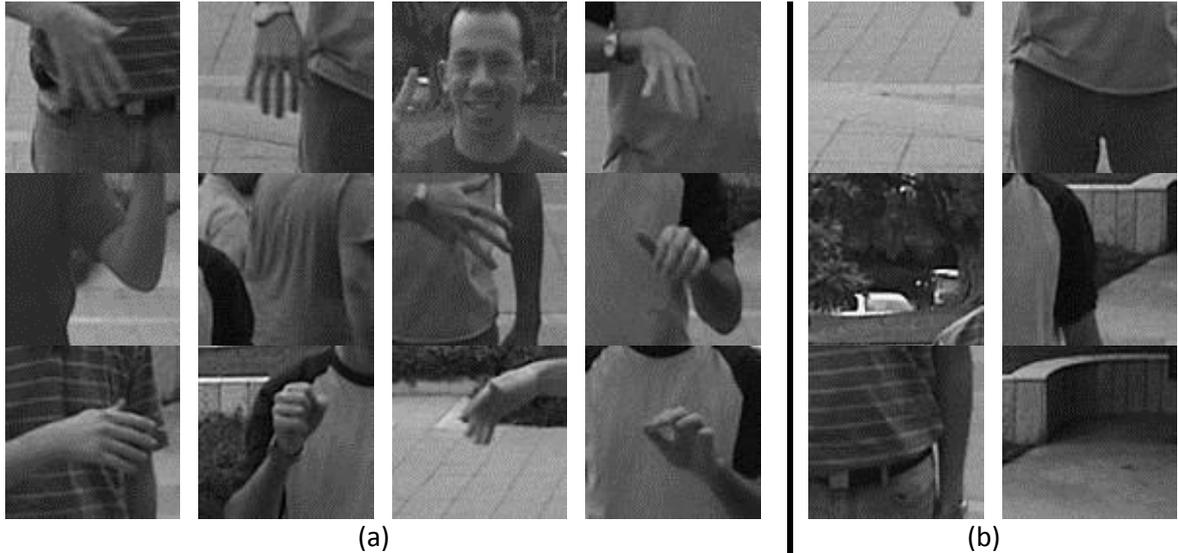

(a) (b)

Figure 32: Image patch examples used for training in appearance learning stage. (a) Positive examples; (b) Negative background examples.

### 4.3.4 Context-based learning stage

In this learning stage hands are considered in the context of the surrounding body parts, and specifically in the context of faces. We utilize the Viola & Jones state-of-the-art face detector (Bradski, 2000), combined with a skin-color detector (Conaire *et al.*, 2007) to detect the face location in each frame of our training video sets.

Given a weak hand detector, we utilize its highest detections when applied to our training video datasets, together with the detected face locations, to learn a context-based hand detector. We utilize a context-based model denoted as the 'chains' model (Karlinsky *et al.*, 2010), in which the relation between context features and the target part is modeled in a non-parametric manner, using an ensemble of feature chains leading from reference parts in the context to the detection target. This model is able to represent highly deformable objects such as the human body for the detection of object parts such as the hands (see Appendix A - The chains model for detecting parts by their context).

The training data consists of whole video frames from the training video sets, in which the weak hand detector detects hands with high confidence levels. We use visual features extracted from the video frames together with spatial positions of both the detected hands and faces during the training process (Figure 33).

As in the appearance learning stage, the training data is noisy, since the detections of both the weak hand detector and the face detector consist of true and false positive examples. No external labeling of hand positions is used during this learning stage.

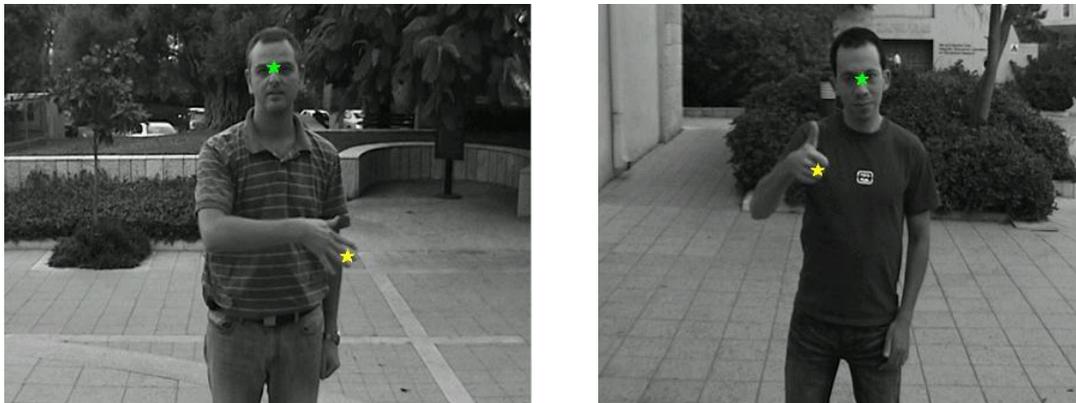

Figure 33: Training examples for the context-based learning stage. The training data consists of a video frame image and the positions of a hand example (yellow stars) and its reference face (green stars), as provided by a weak hand detector and a face detector.

### 4.3.5 Combining appearance and context

Once a new appearance model and a new context-based model are learned, we apply them to the training video sets and combine their detections. For simplicity, we assume that the two models are statistically independent and therefore, we combine their detection voting maps in a simple arithmetic manner (after normalizing each voting map). The combined voting map represents the detection confidence level of a combined appearance-context hand detector (Figure 34).

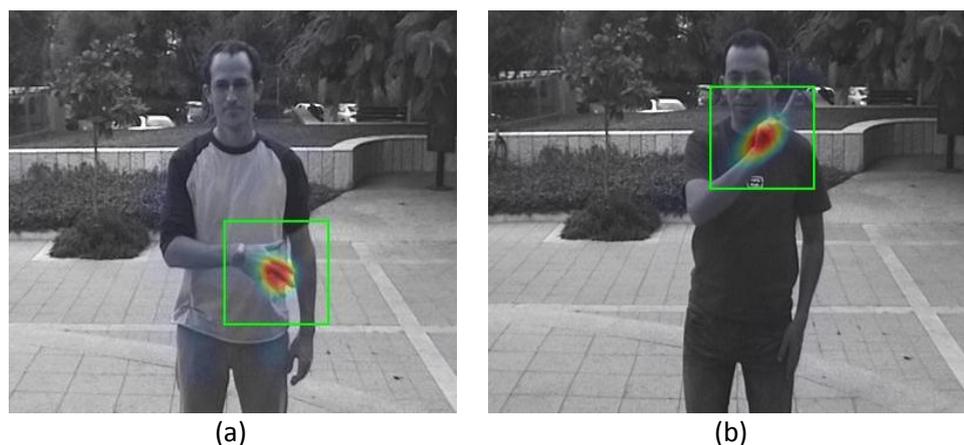

(a)            (b)

Figure 34: Visualization of hand detection confidence maps. The detection confidence temperature map is overlaid on query images (red indicates high confidence level). The greed rectangles indicate image patches that are extracted around detected hand locations and can be used as positive examples for further learning. (a) A 'reaching' hand gesture can be detected based on typical 'mover' appearance; (b) A 'pointing' hand gesture has a very ambiguous appearance and thus can be detected only by combining contextual body information.

**4.3.6 Iterative incremental learning**

The combined appearance-context hand detector yields improved recognition capabilities. It is a more robust detector that accounts for a wider variety of hand poses than the weak detector that was used during training.

We iteratively reapply the above learning procedure to the training video data. At each iteration phase we use the resulting hand detector from the previous phase as the weak detector and apply it to the training data. Next, we use the detections with high confidence scores as positive hand examples for training a new combined appearance-context hand detector. This iterative procedure provides the two learning stages (appearance and context-based) at each iteration, with less noisy labeled examples of a wider range of hand appearances and poses. It should be noted that in actual cognitive learning the co-training process will be continuously applied to new visual data. Our testing on such a learning process yield with similar results.

**4.3.7 Spatiotemporal consistency (object tracking)**

In conjunction with the iterative learning procedure we utilize the spatiotemporal consistency of the video data in order to enrich hand poses and gestures in the training examples. At each iteration phase, after applying the combined hand detector from the previous phase, we track hand detections that have high confidence scores along the video sequences. The spatiotemporal consistency of the video data allows us to extract many additional hand examples at new observed poses and gestures even when initiating from very few hand detections.

We apply a rather naïve tracking approach which independently tracks an image region between every two consecutive video frames. We initiate the tracking process at the frame and position of a detected hand with high confidence score.

First, we refer to a rectangular region (30x30 pixels) surrounding the detection position as the object to be tracked. Next, we describe the object as a set of SIFT features extracted around boundary points of the object. For each feature we also measure its offset from the object center. At the consecutive video frame we search for similar features within a fixed neighborhood around the previous object location. We weigh the different features by their optical flow magnitude (Black and Anandan, 1996) in order to discard of stationary background features. Finally, each of the detected features votes for the object center, which is determined by the maximum sum of votes. This object center is then used to define a new rectangular object region to be tracked into the next consecutive video frame.

Due to the naïve nature of this tracking procedure we limited the tracking period to 50 frames (2 seconds) in order to avoid false tracking trajectories.

## 4.4 Results

We apply our hand recognition learning scheme to a set of unlabeled training video sequences. First, an initial appearance-based model is learned from temporally extended 'mover' events (via tracking). For this learning phase we use the unlabeled *Movers* training video dataset. We proceed with an iterative learning process in which a combined appearance-context hand detector is trained based on positive examples, which are extracted from the most confident detections of the resulting detector from the previous iteration. This iterative process is conducted on the unlabeled *Freely-moving hands* video dataset. In our experiment we iterate this learning process for 3 times (performing additional iterations did not improve much the detection capabilities).

Assuming that the detection capabilities are improved at each iteration, we use an increasing portion of the most confident detections of the preceding detector, as positive examples for training. In our experiment we use the portions 2%, 10% and 20% for the first, second and third iterations respectively.

Images of hands detected with the final combined appearance-context detector are shown in Figure 35. Some difficult hand instances of poor visibility due to motion blur, low contrast and cluttered background could not be detected only by their appearance, but were successfully detected when combined with the context-based model (Figure 35b).

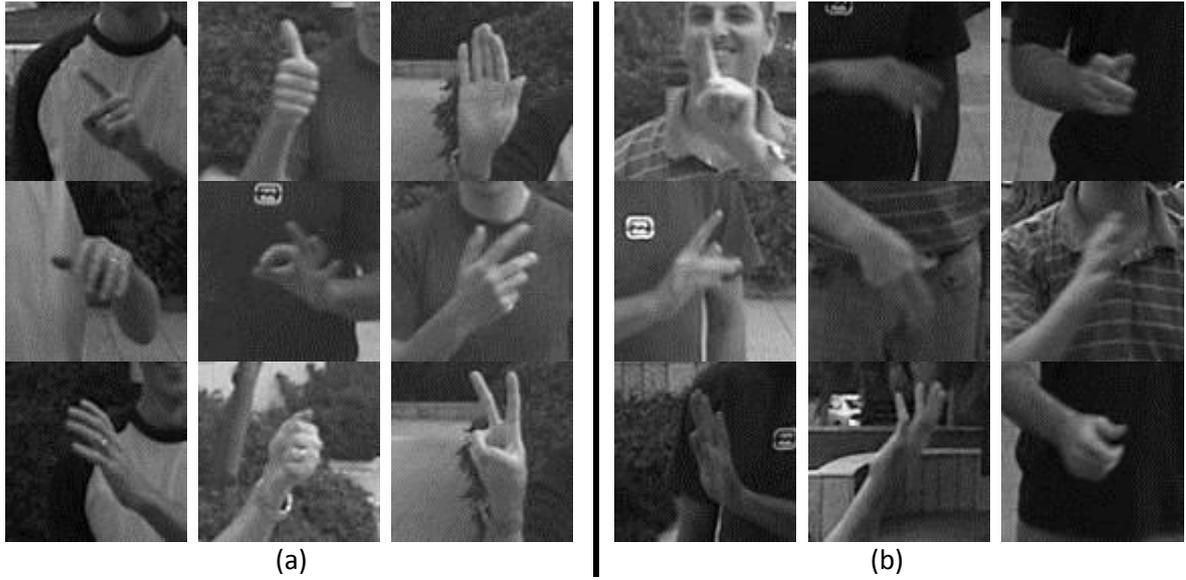

Figure 35: Example images of detected hands. (a) A variety of appearances captured by the final learned hand detector. (b) Some difficult hand poses that were captured by the final combined appearance-context hand detector, but were not captured only by appearance.

To demonstrate the increasing recognition capabilities of our learning scheme, we analyze the detection performance on all left-out video sequences, and create average precision-recall graphs at each learning iteration. We compare the performance of our learned hand detectors to a similar detector trained on manual annotations of the video sequences. Figure 36 shows the increase in recognition capabilities of our learned models, which almost reach the best possible performance provided by the supervised learning process.

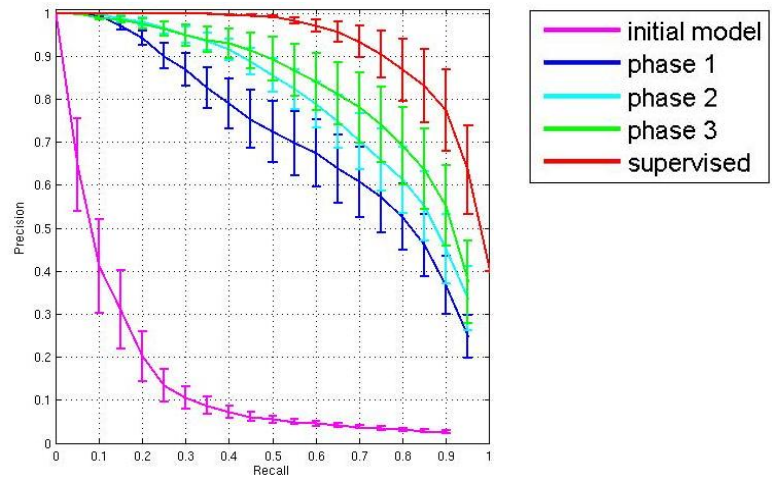

Figure 36: Average precision-recall graph of the combined appearance-context hand detectors learned at each iteration phase of our full learning scheme. The red curve indicates the performance of a hand detector trained in a supervised manner on manual annotations of the training videos.

Since hand poses are usually associated with typical body poses, we measure the spatial coverage of the detections' positions at each learning iteration phase relative the overall spatial coverage of all true hand positions (manually annotated for all video frames). We choose to analyze detections that yield 90% precision in all the learning phases. The increasing spatial coverage demonstrates the increasing range of recognized hand poses and gestures along the iterative learning process (Figure 37).

It is interesting to analyze separately the performance of the appearance-based and context-based detectors at each iteration phase. In Figure 38 we can see that the context-based model yields with a larger increase in performance after the initial phase, while the increase in performance of the appearance-based model is more subtle. At any iteration phase the combined appearance-context based detector yields with a superior performance than the two individual models.

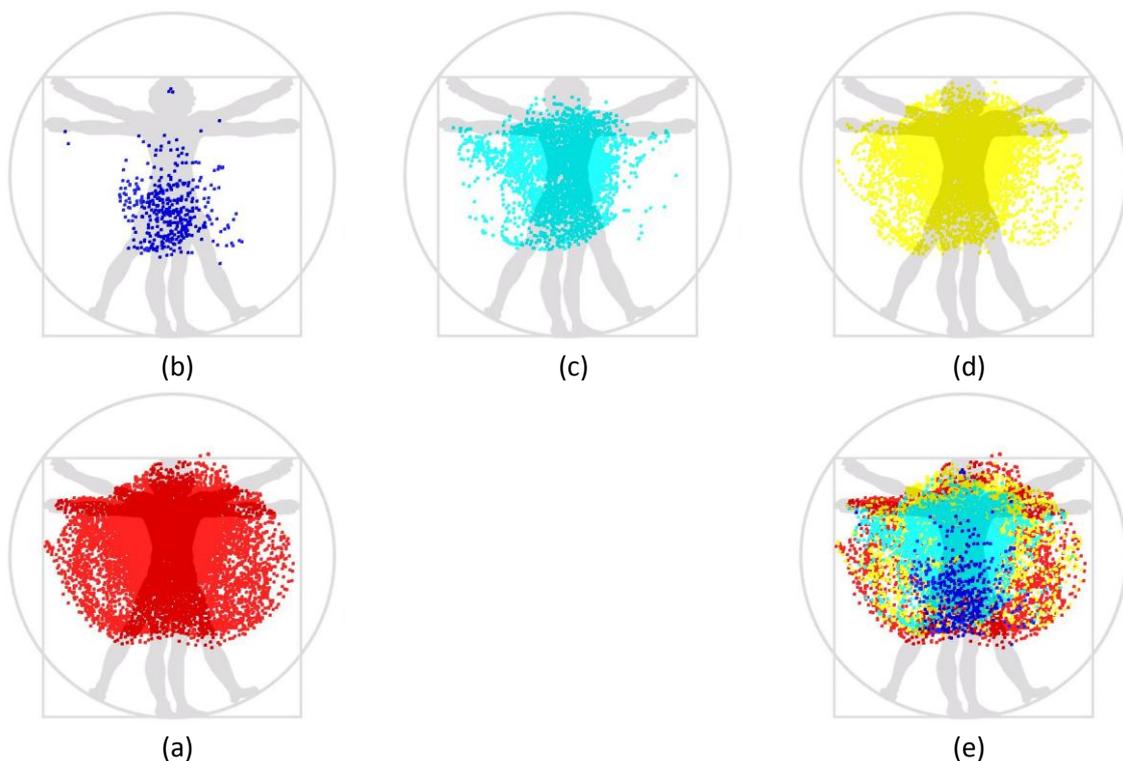

Figure 37: Spatial distribution of detected hand positions at different learning phases. (a) 100% spatial coverage of all the true hand positions; (b) Initial model - 22% spatial coverage; (c) Phase 1 – 55% spatial coverage; (d) Phase 3 – 81% spatial coverage; (e) An overlay of all phases on-top of the ground truth.

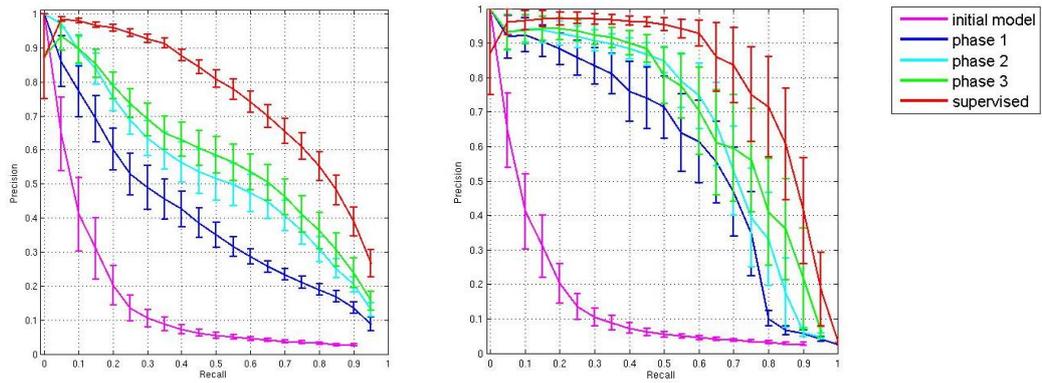

Figure 38: Performance analysis of the two learning stages: (a) Appearance-based detectors; (b) context-based detectors. The red curve indicates the performance of the relevant supervised learning process.

Object tracking and body context have a major role in our hand learning scheme as they provide important sources of information. To illustrate the significance of these sources of information, we have conducted a second experiment, while applying two modified learning schemes: 'No-Tracking' scheme in which no object tracking is used to extend detection results along temporal trajectories; and 'No-Context' scheme in which only an appearance-based detector is learned at each iteration phase (without context). Figure 39 presents the average precision-recall graphs of the resulting detectors at each learning iteration phase for the two learning schemes. The graphs show a significant increase in performance only in the first iteration phase, while additional phases do not provide increasing recognition capabilities.

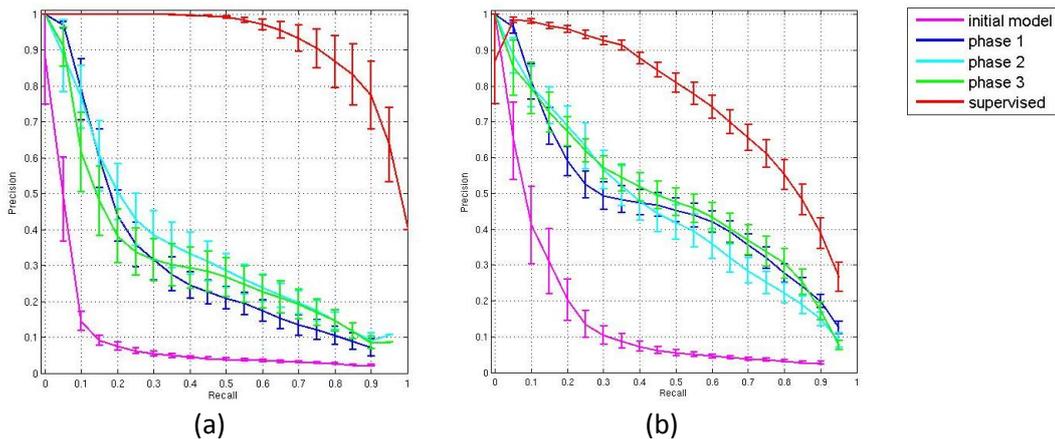

(a)          (b)

Figure 39: Average precision-recall of learned detectors in: (a) 'No-Tracking' scheme; (b) 'No-Context' scheme.

Figure 40 presents a performance comparison for the final detectors learned with the three different learning schemes. Our complete learning scheme yields significantly superior recognition capabilities, which verifies the importance of both object tracking and context in the overall learning process of hands.

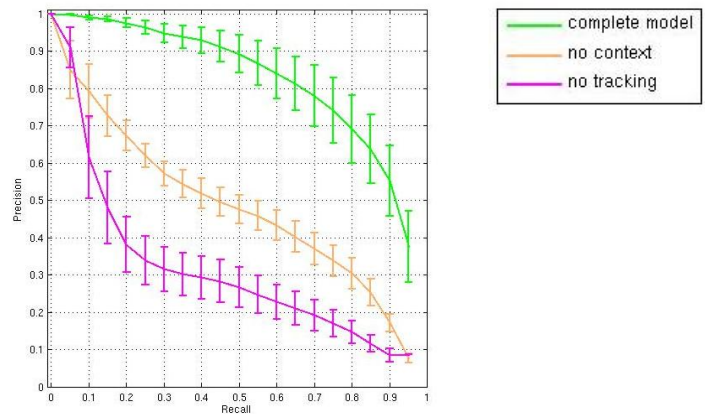

Figure 40: Performance comparison of final hand detectors learned using our full scheme (green curve), 'No-Tracking' scheme (magenta curve), and 'No-Context' scheme (orange curve).

## 4.5 Summary

The recall rate of the 'mover'-based hand detector that was introduced in chapter 3 is initially low, since detection is limited to specific hand configurations extracted as 'mover' events, typically engaged in object grasping. In this chapter we show how recall rate rapidly increases in subsequent iterations of the learning algorithm based on two mechanisms: tracking and body context. Detected hands are tracked over a short period and additional hand images are extracted by the end of the tracking period. Context is useful for detecting hands based on surrounding body parts in ambiguous images. We therefore included in the model an existing algorithm that uses surrounding body parts, including the face, for hand detection (Appendix A - The chains model for detecting parts by their context).

As hands are detected based on either appearance or body context, we found that the two detection methods cooperate to extend the range of appearances and poses used by the algorithm. The context-based detection successfully recognizes hands with novel appearances, provided that the pose is already known ('pose' here is the configuration of context features, on the shoulders, arms, etc.). The newly learned appearances lead in turn to the learning of additional poses. The learning was applied to the input videos in several iterations. The results show that appearance-based learning and context-based learning guide each other to boost recognition. Performance improves rapidly during the first 3 iterations, approaching the performance of fully supervised training on the same data.

The suggested algorithm focuses on motion-based cues, but additional visual mechanisms (e.g., biomechanical motion (Bertenthal *et al.*, 1985)) as well as non-visual sensory motor (Sommerville *et al.*, 2005), supplied in part by the mirroring system (Rizzolatti and Sinigaglia, 2008), may also play a role in the learning process. In the next chapter we explore, a possible contribution to the learning process of hands from observing one's own hands in motion.

# 5 First person perspective of own hands

## 5.1 Overview

In chapter 3 and chapter 4 we have suggested an algorithm for learning to detect hands across a broad range of appearances and poses using natural dynamic visual input and without external supervision. The algorithm is based on statistical learning of repeating configurations in the input, combined with innate structures in the form of simple specific 'proto concepts'. These include the 'tagging' of 'mover' events, the use of spatiotemporal continuity in tracking, and the automatic association between 'mover' events and face features. The algorithm focuses on motion-based cues, but additional visual as well as non-visual sensory motor mechanisms (Piaget, 1952; Rizzolatti and Sinigaglia, 2008), may also play a role in the learning process. In particular, a possible source of information for learning about hands can be obtained by an infant by moving and observing its own hands. Both behavioral (Sommerville *et al.*, 2005) and physiological evidence support the use and representation of 'own' hands, but their possible role in developing hand detection remains unclear.

In this chapter, we evaluate the use of 'own' hand images to train a hand classifier. First person perspective images of own hands were obtained from two adult subjects using video cameras placed close to the subject's head (roughly similar to (Yoshida and Smith, 2008), see Figure 41). These training images allow us to evaluate the limitations of 'own' hand images under favorable training conditions (good imaging conditions and a broad range of hand configurations).

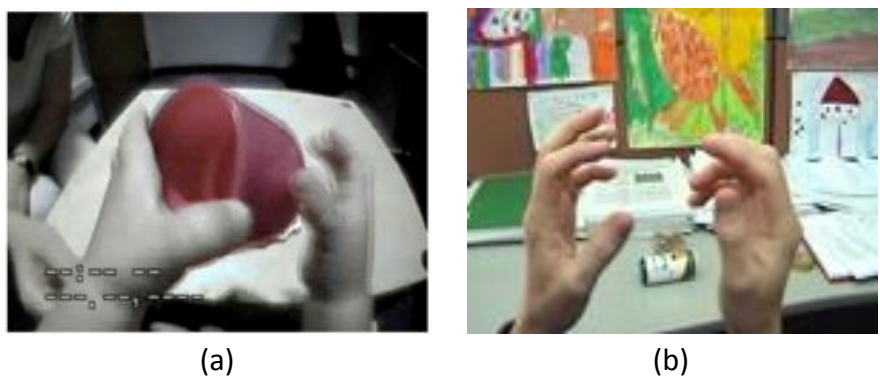

(a) (b)

Figure 41: Sample images of first person perspective of own hands: (a) image from Yoshida & Smith (Yoshida and Smith, 2008); (b) image from our dataset.

Our testing shows, that using images of 'own' hands is not as effective as using 'mover' instances in detecting hands in general, and in detecting hands engaged in object manipulation in particular.

## 5.2 Data

The training data consists of 2 video sequences taken with a static video camera at a fixed exposure setting. The video sequences, about 120 seconds each, were converted to gray level. Each video frame is down-sampled to a 144x112 pixels resolution, to roughly match a convenient hands scale.

Each video sequence depicts a human perspective of his own hands while moving them around and manipulating some objects on a table in front of him. For this purpose, the camera was located at a fixed position adjacent to the subject's head, in parallel direction to the subject's field of view, thus imitating the subject's own perspective. We used the same static background for both videos.

The test data consists of 2 video datasets (Figure 16 and Figure 17) with a total of 16 outdoor video scenes (8 scenes per dataset), taken with a static video camera at a fixed exposure setting. The video sequences, about 60 seconds each, were converted to gray level. Each video frame is at a 436x336 pixels resolution. We refer to these datasets as *Freely-moving hands* dataset and *Manipulating hands* dataset (see section 3.5.1 for more details).

## 5.3 Results

Our experiment's training phase consisted of learning two different hand detectors from each of the training video sequences. From each video sequences, we first utilize optical flow maps (Black and Anandan, 1996) to extract image patches containing large moving parts. These image patches mostly contain hands (Figure 42), but may also contain non-hands images due to shadows and reflections (background was static). We then collected non-hands image patches extracted from non-person images of the PASCAL VOC2007 dataset.

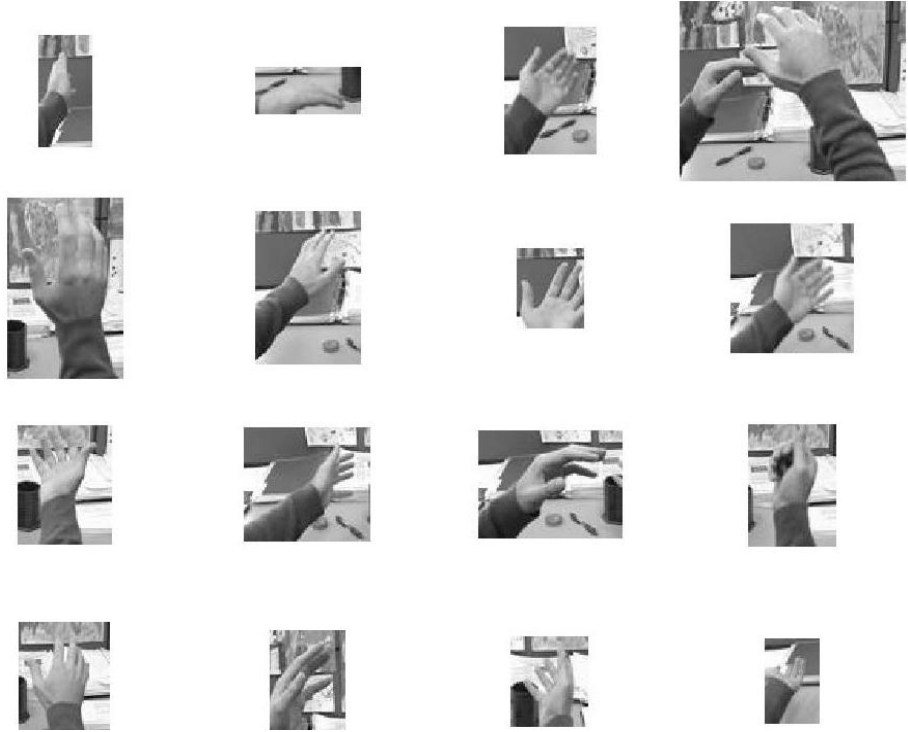

Figure 42: Own hands training. Extracted image patches based on large moving parts.

We used the noisy hands patches as positive examples and the non-hands patches as negative examples, for learning two appearance-based hand detectors (one per training video) based on the ANN-star model (Karlinsky *et al.*, 2010), in a supervised manner.

In the testing phase of our experiment we applied the two hand detectors to the two test datasets, and compare their detection performance with our 'mover'-based hand detector (Figure 43).

## 5.4 Discussion

The results show that the 'own' hands detector did not generalize to the *Manipulating hands* dataset (maximal precision less that 5%). For *Freely-moving hands*, in the low-recall (2%), high-precision range (which is relevant for subsequent training), 'own' hands based detector reached about 15% compared with 97% of the 'mover' based detector. It is interesting to note in this regard that data from infants' behavior indicates that their looking time is high particularly for hands engaged in object manipulation (Aslin, 2009; Frank *et al.*, 2011). This finding is more consistent with computational results obtained from 'mover' based training compared with 'own' hands training, probably because generalization is more straightforward.

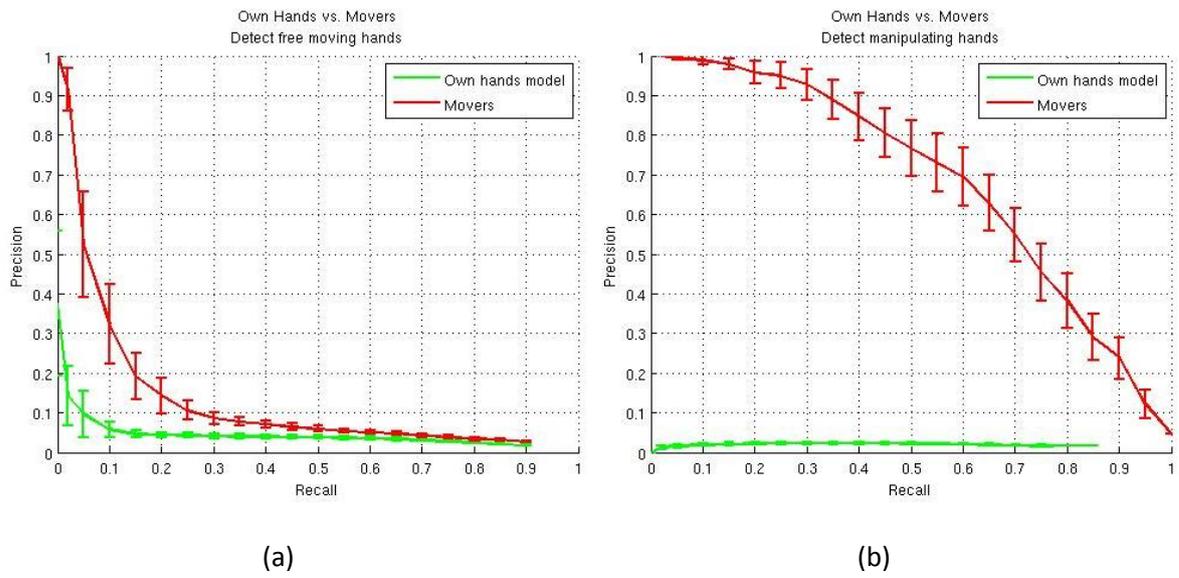

Figure 43: Average precision-recall of hand detectors trained on 'own' hands and detected 'mover' events. (a) *Manipulating hands* dataset; (b) *Freely-moving hands* dataset.

# 6 Learning direction of gaze from 'mover' events and faces

## 6.1 Overview

In chapter 3 and chapter 4 we have described an algorithm which demonstrates the unsupervised learning of human hands in complex natural scenes. The algorithm is guided by simple 'proto-concepts' ('mover' in our example) leading to the automatic acquisition of hands, which do not emerge automatically without domain-specific biases.

In this chapter we propose another example which demonstrates this guidance mechanism. In particular, we propose that 'mover' events can provide accurate teaching cues in the acquisition of another intriguing capacity in early visual perception – detecting direction of gaze, i.e. following another person's gaze based on head orientation and later eyes direction (Scaife, 1975; Flom and Lee, 2007). Developmental research indicates that this skill, which starts at about 3-6 months and continues to develop, plays an important role in the development of communication and language (Tomasello, 1999). However, it remains unclear how this learning might be accomplished since cues for direction of gaze (head and eyes) can be subtle and difficult to extract and use (Murphy-Chutorian and Trivedi, 2009).

## 6.2 Data

We used 8 movies, each depicting a different subject picking up objects from a table and putting them back down (Figure 44). Each movie contained around 50 pairs of pick up and put interactions.

We manually labeled all 887 events of object interactions. For each object interaction we marked the center of the target object at the frame of initial contact, and the center of the face. The direction from the face-center to the target is used as ground truth for measuring performance.

## 6.3 Automatic annotation of gaze direction

We used our 'mover' detector to detect object interaction events. We consider the duration of each event to be between the first incoming motion and the last outgoing motion. In these movies, subjects typically look at an object, reach towards it, and then look towards the next location (putting the object down) even before making contact with the object. The time lag between looking at the object and making contact with it is somewhat variable. We assumed that 10 frames (0.4 seconds) before the middle of the event, gaze is directed towards the event location – this is usually true, but not always.

In each frame of interest, the face is detected using Viola-Jones face detector from OpenCV together with a skin color detector (Conaire *et al.*, 2007). We take the direction from the center of the face to the center of the cell that triggered the event as the gaze direction (Figure 47). If a face is not detected we ignore this event.

Note that these annotations are noisy. The object interaction event may be wrong, face detection may be wrong, the centers of both face and target are not accurate, and sometimes the gaze is not directed towards the target.

## 6.4 Nearest Neighbors model

Our goal is to detect gaze direction in the test frames. We use a leave-one-out scheme. In each repetition we train on the automatic annotations of 7 movies and test on the test frames of the left out movie.

For each training event we extract a region around the face, and compute a HOG descriptor (Dalal and Triggs, 2005). The HOG descriptors are calculated using the code supplied by the authors with default parameter values. We associate this descriptor with the gaze direction of the event.

For each test frame, we first detect the face center (see above). We are not interested in evaluating the face detector, so if we fail to detect the face (no detection or detection was not within 30 pixels of the ground truth), we discard the test image. This left 96% of the original test images.

We then compute a HOG descriptor for the detected face and find its 10 nearest neighbors in the training data (other movies), using L2 norm. The predicted gaze direction is a weighted average of the gaze directions of the neighbors (weighted by similarity of HOG).

## 6.5 Human control

We compared our results with human performance on the same test images. We showed two human subjects the same face patches that were used by the algorithm. These are the patches of automatically detected faces, at the beginning of each object interaction in the test movie. We asked each subject to mark the gaze direction of each face patch.

## 6.6 Results

For the purpose of analysis we consider a prediction to be correct if it falls within 20 degrees of the ground truth. This threshold leads to average accuracy of 75% by the human subjects. Note that the ground truth was marked according to target objects, so in the few cases where the actor is looking elsewhere, the ground truth is wrong.

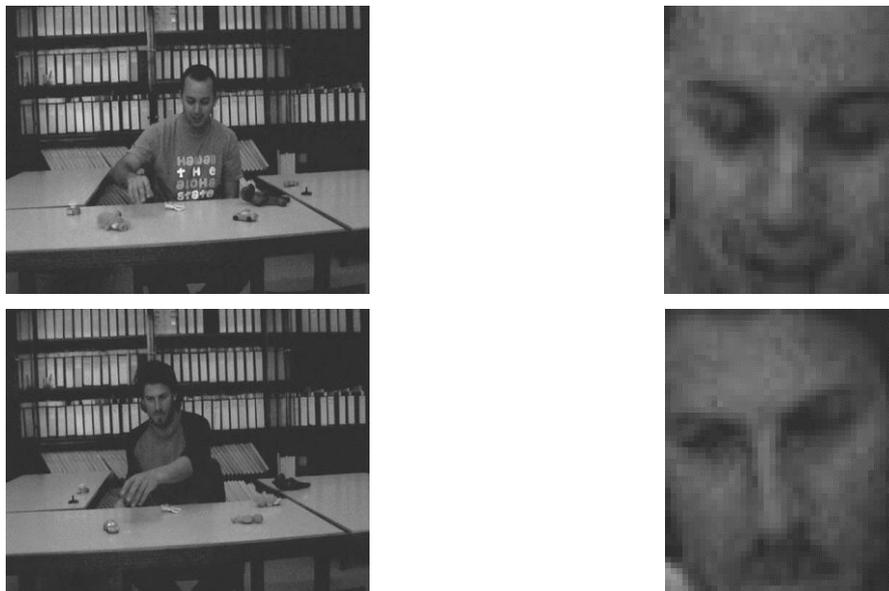

Figure 44: Sample images from the 'Gaze' video dataset. Left: A few images from our data. Right: Zoom in on the face region, showing the low resolution.

The results are given in Table 7 and Figure 45. Averages are over the 8 different movies. Each movie contains around 100 test frames. Figure 46 gives a more complete picture of the performance for all possible error thresholds. Visualization of detected direction of gaze compared to human performance is shown in Figure 48.

| Model | Average % correct | Standard Deviation |
|---|---|---|
| Nearest Neighbors | 71 | 14 |
| Chance guess (according to distribution of train data) | 34 | 5 |
| Human subject 1 | 70 | 13 |
| Human subject 2 | 81 | 13 |

Table 7: Performance of algorithm vs. human and chance

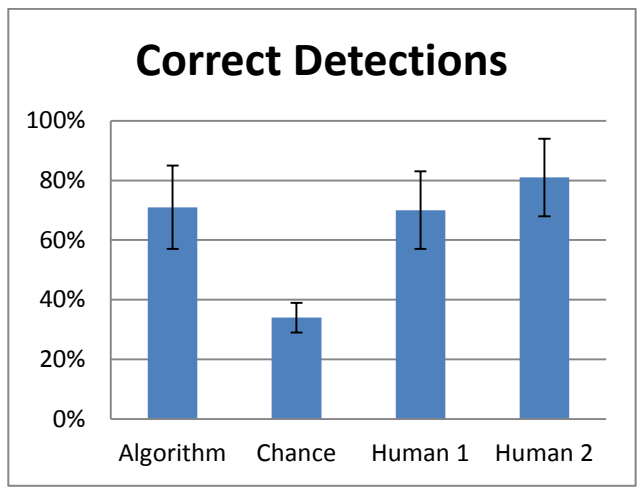

Figure 45: Performance of algorithm vs. human and chance

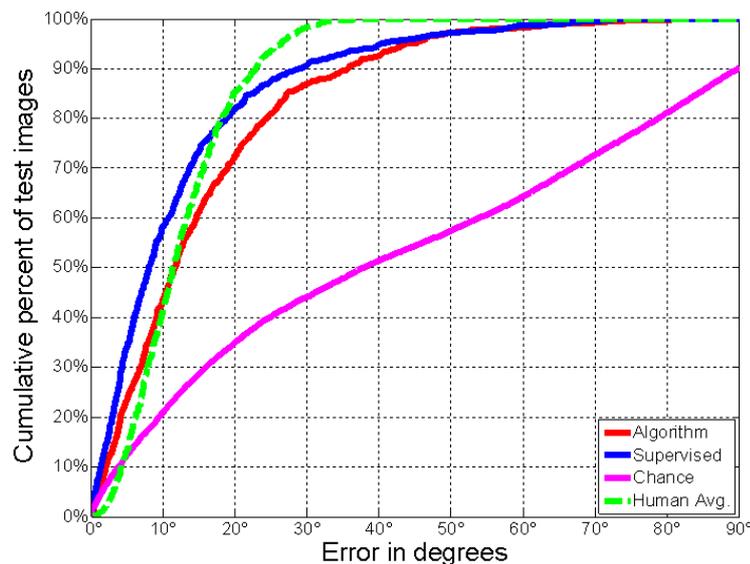

Figure 46 : Performance of gaze detectors: detector trained on mover events (red), detector trained with full manual supervision (blue), chance level (magenta) and human performance (dashed green). Chance level was computed using all possible draws from the distribution of gaze directions in the test images. Human performance shows the average performance of two human observers. Abscissa: maximum error in degrees. Ordinate: Cumulative percent of images.

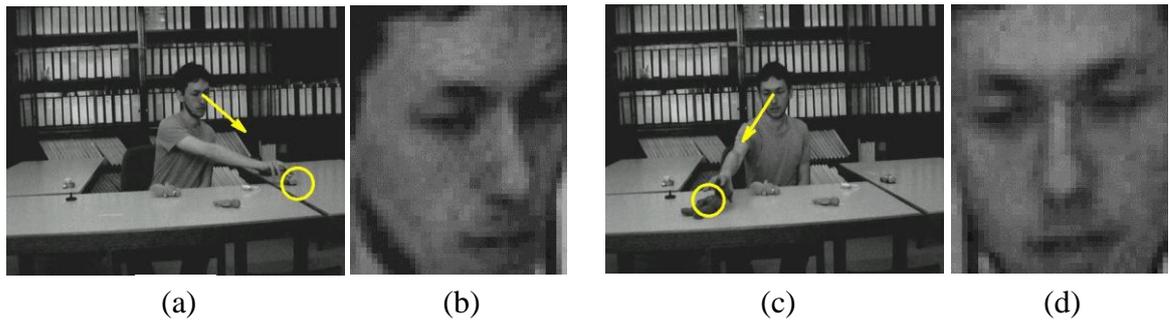

(a) (b) (c) (d)

Figure 47: Automatic annotations used for training the gaze direction detector. (a+c) Detected 'mover' event. Yellow circle marks location of event, providing teaching signals for gaze direction (yellow arrow). (b+d) Face image region used for gaze learning.

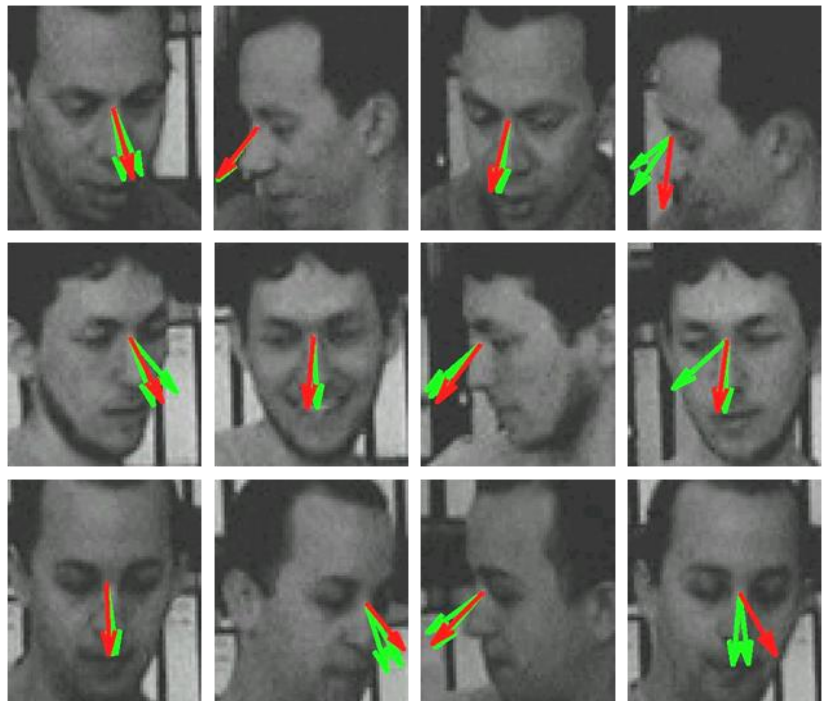

Figure 48: Predicted direction of gaze: results of algorithm (red arrows) and two human observers (green arrows).

## 6.7 Summary

Infants' looking is attracted by other people's hands engaged in object manipulation (Aslin, 2009; Frank *et al.*, 2011) and they often shift their gaze from face to a manipulating hand (Amano *et al.*, 2004). People often look at objects they manipulate, especially slightly before and during initial object contact.

In this chapter we have proposed an algorithm which uses 'mover' events (see chapter 3) as an internal teaching signal for learning direction of gaze based on head orientation and eyes direction. Our algorithm detects presumed object contacts by detecting

'mover' events, extracts a face image at the onset of each 'mover' event, and learns, by standard classification techniques (k nearest neighbors), to associate the face image with the 2-D direction to the contact event. We assume that initial face-detection is present prior to gaze-learning and locate the face with an available face detector. The resulting classifier estimates gaze direction in novel images of new persons with accuracy approaching adult human performance under similar conditions

# 7 Conclusions

## 7.1 Summary

In this thesis we have described the use of visual motion information and internal supervision in the task of object recognition. We have developed an adaptive part-detection model for a dynamic environment. Using motion and spatiotemporal consistency, our model adapts online to dynamic changes of an object, by gradually updating the appearance and structure of the object and its parts. The model combines the dynamics of the visual input to yield with a powerful realistic recognition system, with capabilities beyond object tracking such as the automatic acquisition of object's appearance and structure in 3-D space.

We have also developed methods for the use of image motion for the notably difficult tasks of recognizing human body parts (in particular hands) and following direction of gaze. Using natural dynamic visual input and without external supervision, our model learns to detect hands across a broad range of appearances and poses, and to extract direction of gaze. Our model uses image motion as an internal teaching signal via the detection of active-motion, we call 'mover' events, to guide the learning system along a path, which leads to the gradual acquisition of complex concepts. Our model also uses the spatiotemporal continuity in tracking, and an association between 'mover' events and face features. We suggest a co-training process of two complimentary detection capabilities (by appearance and context) which guide each other, to boost recognition performance.

We believe that the approach and results open a general study area of 'computational cognitive development', with a specific, surprising and potentially controversial result. Learning body-parts detection and gaze direction are two capacities in which the gap between computational difficulty and infant learning is particularly striking. To our knowledge, our algorithm is the first demonstration of learning to detect hands and gaze direction in an unsupervised manner, in natural images. The results offer a novel general approach to the combination of learning and innate mechanisms in human cognition, and demonstrate its power with striking examples. We show that meaningful complex concepts acquired in development may be neither learned on their own nor innate. Instead, simple domain-specific internal teaching signals can guide the learning system along a path, which leads to the gradual acquisition of complex concepts that do not emerge automatically otherwise. (See Appendix C – List of publications.)

## 7.2 Key contributions

- **Adaptive part detection model for a dynamic environment**. Our approach is a natural extension of the existing part-detection methods for static images. We combine two sources of information in constructing the object model at time $(t + \Delta t)$: compatibility with the measured optical flow between time frame t and $(t + \Delta t)$, and similarity to the object model at time t. Our scheme dynamically updates the model parameters from frame to frame, utilizing the interpreted object appearance and geometry. We suggest a simple new way of online updating the object model by adaptive approximate nearest neighbors search and statistical kernel density estimation. We demonstrate how our model adaption can also be used for learning with no external supervision, by extending initial recognition capabilities to a new set of viewing directions.

- **Using different types of visual motion.** Existing methods use visual motion for segmentation, separating moving regions from a stationary background, and tracking a moving region. However, our simulations showed that general motion cues on their own are unlikely to provide sufficiently specific cues for learning complex recognition capabilites such as hand detection. The human visual system is sensitive to specific types of visual motion, including launching, active (causing other objects to move), self-propelled, or passive. Thus specific types of visual motion may have different roles in visual processes in general, and in learning recognition capabilities in particular. We suggest the detection of active motion, which we call 'mover' event, as a domain specific teaching signal for learning hands. Other types of motion such as self-propelled and passive motion, may also be used in future research.

- **Learning hands by internal supervision from 'mover' events.** We propose the automatic learning of hands from the detection of active motion, we call 'mover' event, defined as the event of a moving image region causing a stationary region to move or change after contact. 'Mover' detection is simple and primitive, based directly on image motion without requiring object detection or region segmentation. Using natural dynamic visual input and without external supervision our model detects 'mover' events as presumed hand candidates, and uses them to train an appearance-based hand detector. Hands are detected with high accuracy, but are limited to specific hand configurations, since 'mover' events are typically engaged in object grasp and release.

- **Combining appearance and context in co-training.** Context is useful for detecting hands based on surrounding body parts in ambiguous images. We therefore included in our model an existing algorithm that uses surrounding body parts, including the face, for hand detection. As hands are detected based on either appearance or body-context, we found that the two detection methods cooperate to extend the range of appearances and poses used by the algorithm. The context-based detection successfully recognizes hands with novel appearances, provided that the pose is already known. The newly learned appearances lead in turn to the learning of additional poses. Our results show that appearance-based object detection and context-based part-detection guide each other to boost recognition performance.

- **Learning direction of gaze by internal supervision: combining face detection and 'mover' events.** We propose that detected 'mover' events provide accurate teaching cues in detecting and following another person's gaze based on head orientation, and eyes direction. Our algorithm detects presumed object contacts by detecting 'mover' events, extracts a face image at the onset of each 'mover' event, and learns, by standard classification techniques, to associate the face image with the 2-D direction to the contact event. We assume that initial face-detection is present prior to gaze-learning and locate the face with an available face detector. The resulting classifier estimates gaze direction in novel images of new persons with accuracy approaching adult human performance under similar conditions.

## 7.3   Future research

- **Online model updating.** Our adaptive parts detection model for dynamic environments utilizes an online updating scheme of the object model by adaptive approximate nearest neighbors (ANN). The ANN framework is naturally limited in memory, but currently is uncapable of 'forgetting' past data. When applied in an online learning sceme to dynamic visual data, the framework rapidly runs out of memory. It will be interesing to explore online updating techniques for approximate nearest neighbors methods for large scale dynamic input data such as video sequences. Such methods should have an integrated 'forgetting' mechanism that can allow the pruning of irrelevant past data, while allowing new data to be memorized. (One relevant alternative is the hierarchical Dirichlet processes representation.)

- **Computational cognitive development.** Explore other unsupervised learning tasks related to 'computational cognitive development' such as understanding object interactions and recognizing manipulable objects, that can lead to the unsupervised learning of human actions in static images. Manipulable objects may be learned by combining salient image regions and direction of gaze. Object interactions, such as object containment, may be learned by first understanding their ordinal depth through relative motion and motion segmentation.

# Declaration

I hereby declare that this thesis was composed by me, and that I was a major or equal contributor to all the work contained herein, except where stated otherwise. I also declare that this work has not been submitted for any other degree or professional qualification.

# List of Figures

# List of Tables

# Appendix A - The chains model for detecting parts by their context

Addressing the task of recognition, object parts can be ambiguous and are often recognized by the surrounding context of other parts. In particular, the use of context for recognition of non-rigid and deformable objects is more difficult since the context is highly variable. For example, a face in the image can supply a context for hand detection, but the position of the hand is not directly given by the position of the face. We have introduced the chains model (Karlinsky *et al.*, 2010) in which the context is used by forming an ensemble of feature chains that connect a reference part (an easily detectable part such as the face) to a target part (a hard to detect deformable part such as the hand, Figure 49).

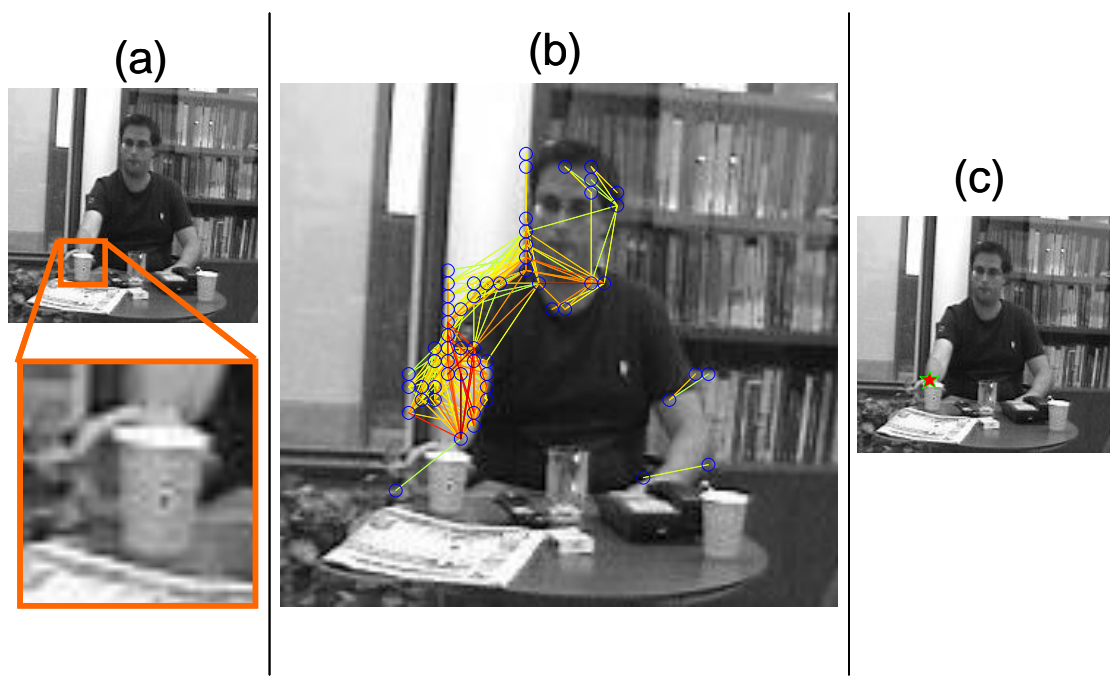

Figure 49: Visualization of the chains model inference. (a) Here the object part of interest is the hand which may not be recognizable by its own appearance; (b) The chains model's inference graph of the query image (the edges are color-coded by their weights); (c) The star indicates the location of the detected hand using disambiguating chains from the face.

This non-parametric model is suitable for specific part detection given a reference part, and also for complete object detection with no initial reference. The method can successfully generalize between different people and backgrounds, and consists of a simple and efficient inference algorithm over the ensemble of possible feature chains. A schematic illustration of the chains model is shown in Figure 50a. For more details on the inference and its implementation see the paper attached in Appendix C – List of publications (Karlinsky *et al.*, 2010).

My contribution in this work is constructing a generic star-like appearance-based object detector (ANN-star) as a special case of the chains model. Similar to part detection, the model can use the test image features to point at the object location, either directly, or via intermediate feature chains (Figure 50b,c).

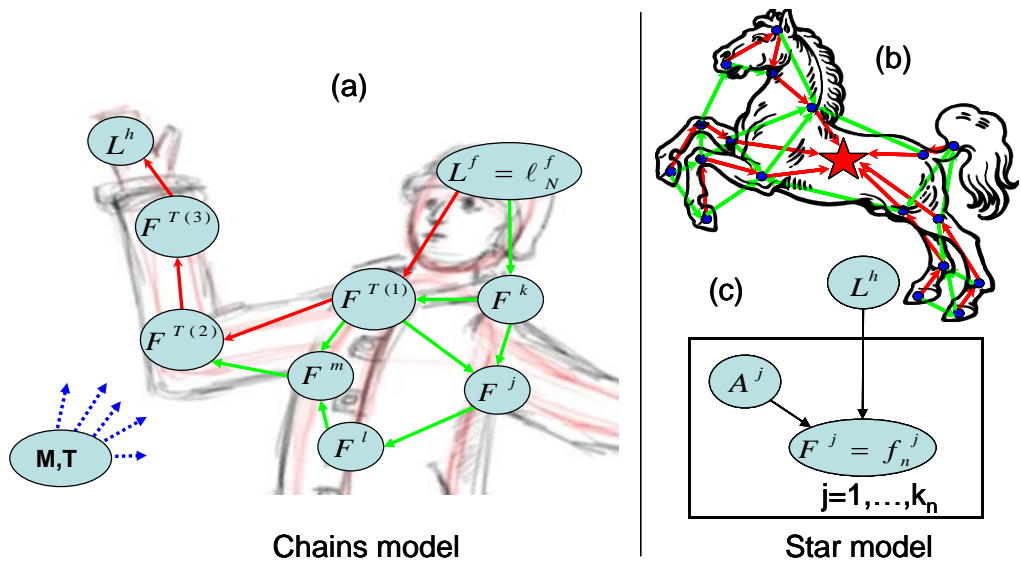

Figure 50: Schematic illustration of the chains graphical models. (a) Chains model applied for part detection. The unobserved variable $L^h$ is the location of the target part (e.g. hand). $L^f$ is the observed location of the reference part (e.g. face). The edges symbolize the chains 'feature graph' constructed over the set of observed features $\{F^j\}$. We consider all simple paths T (red) of length M on the graph. Features not on this path are generated from their 'world' distributions. During inference we marginalize over T and M, summing over paths on the graph going from the reference part to each candidate location of the target part. (b+c) The chains model as an appearance-based object detector (ANN-star).

We use both the appearance-based object-detector and the context-based part-detector as the underlying models for hand detection in chapter 3 and chapter 4 (see Figure 51 for detection examples of the two detectors).

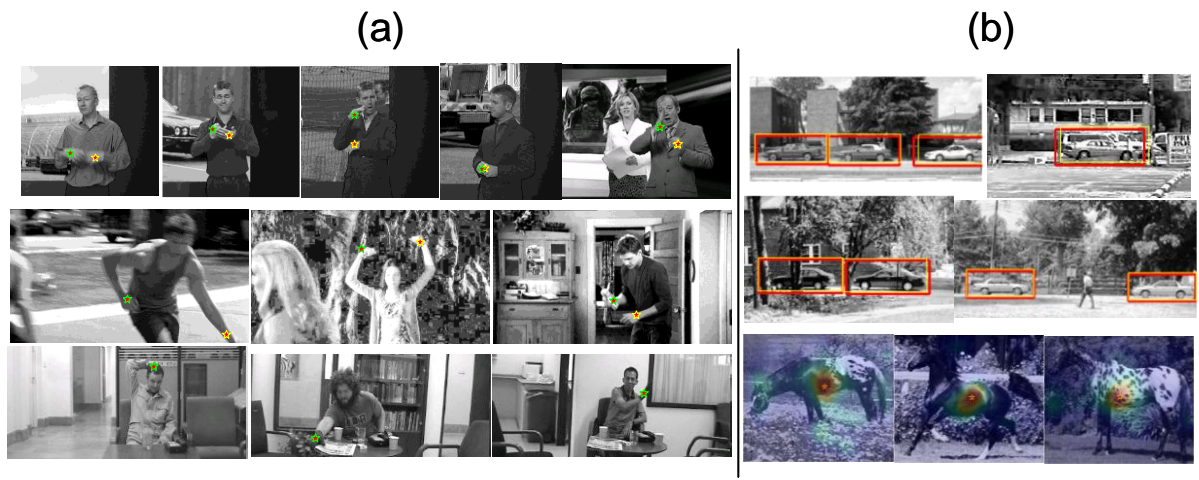

Figure 51: Example of detection results. (a) Hands detection results of the context-based part detector (green and yellow stars indicated first and second maximum respectively); (b) Detection results of the appearance-based object detector for horses and cars.

# Appendix B - Basic-level categorization of intermediate complexity fragments reveals top-down effects of expertise in visual perception

Following the fragment-based object recognition model by Ullman *et al.* (Ullman *et al.*, 2002; Ullman, 2007), it was suggested that objects are represented in the visual cortex by a combination of image-based category-specific pictorial features called 'fragments'. In a previous study, based on EEG recordings from human subjects, it was demonstrated that the abstract computational measure of the mutual information (MI) between the fragments and the category of images is psychologically real with neurophysiological consequences (Harel *et al.*, 2007).

I have collaborated with a team from the department of psychology at the Hebrew university, on a research examining how visual expertise[1] affects basic-level categorization of intermediate complexity (IC) fragments (Ullman *et al.*, 2002), which do not provide a holistic image of the object and cannot be processed with respect to object's configuration (Harel *et al.*, 2011). In the experimental procedure of this work, car experts and novices categorized computer-selected image fragments of cars, airplanes, and faces. Within each category, the fragments varied in their MI, which is used as an objective quantifiable measure of feature diagnosticity.

My contribution to this research was to provide the sets of category-specific image fragments of various sizes as the visual stimuli for the experiments, as well as the corresponding MI measures of these image fragments. Overlapping image fragments were pruned such that large fragments containing smaller fragments with higher MI were discarded.

The findings show that the categorization of face and airplane fragments was similar within and between groups, showing better performance with increasing MI levels. Novices categorized car fragments more slowly than face and airplane fragments, while experts categorized car fragments as fast as face and airplane fragments. The difference between car experts and novices was not in the way they categorized IC car fragments, but in the overall faster reaction times of the car experts. Though the fragments by themselves were equally informative to the experts as they were to the novices, the experts overcame the need for additional processing time required for the novice to evaluate the perceptual evidence and reach a decision. This might be achieved by applying top-down mechanisms (such as experience-based knowledge and attention) that allocate larger resources to fragments from the category of expertise, leading to their faster categorization. For more details see the paper in Appendix C – List of publications (Harel *et al.*, 2011).

---
[1] Visual expertise is usually defined as the superior ability to distinguish between exemplars of a homogeneous category.

# Appendix C – List of publications

1. **The chains model for detecting parts by their context**
   L. Karlinsky, M. Dinerstein, D. Harari, S. Ullman.

   *Proceedings of Computer Vision and Pattern Recognition* (2010)

2. **Basic-level categorization of intermediate complexity fragments reveals top-down effects of expertise**
   A. Harel, S. Ullman, D. Harari, S. Bentin.

   *Journal of Vision* **11**(8):18, 1–13 (2011)

3. **From simple innate biases to complex visual concepts**
   S. Ullman, D. Harari and N. Dorfman.

   *Proceedings of the National Academy of Sciences* **109**(44):18215-18220 (2012)

4. **Extending Recognition in a Changing Environment**
   D. Harari and S. Ullman.

   *Proceedings of the International Conference on Computer Vision Theory and Applications* **1**:632-640 (2013)